\pdfoutput=1

\documentclass[11pt]{article}

\usepackage[preprint]{acl}

\usepackage{times}
\usepackage{latexsym}
\usepackage{arydshln}
\usepackage{tikz}
\usepackage{placeins}

\usepackage[T1]{fontenc}

\usepackage[utf8]{inputenc}

\usepackage{microtype}

\usepackage{inconsolata}

\usepackage{graphicx}
\usepackage{booktabs}
\usepackage{multirow}
\usepackage{enumitem}
\usepackage{subcaption}

\title{A Post-trainer's Guide to Multilingual Training Data: \\Uncovering Cross-lingual Transfer Dynamics}

\author{
  \textbf{Luísa Shimabucoro\dag\textsuperscript{1,2}},
  \textbf{Ahmet Üstün\textsuperscript{1}},\\
  \textbf{Marzieh Fadaee\textsuperscript{1}},
  \textbf{Sebastian Ruder\textsuperscript{3}}
\\
\\
  \textsuperscript{1}Cohere For AI
  \textsuperscript{2}University of São Paulo
  \textsuperscript{3}Meta
}

\begin{document}

\maketitle

\def\thefootnote{\dag}\footnotetext{Corresponding author: \texttt{luisashimabucoro@gmail.com}}

\begin{abstract}
In order for large language models to be useful across the globe, they are fine-tuned to follow instructions on multilingual data. Despite the ubiquity of such post-training, a clear understanding of the dynamics that enable cross-lingual transfer remains elusive.
This study examines cross-lingual transfer (CLT) dynamics in realistic post-training settings. We study two model families of up to 35B parameters in size trained on carefully controlled mixtures of multilingual data on three generative tasks with varying levels of complexity (summarization, instruction following, and mathematical reasoning) in both single-task and multi-task instruction tuning settings. Overall, we find that the dynamics of cross-lingual transfer and multilingual performance cannot be explained by isolated variables, varying depending on the combination of post-training settings. Finally, we identify the conditions that lead to effective cross-lingual transfer in practice.

\end{abstract}

\section{Introduction}

\begin{figure}[ht!]
  \centering
  \includegraphics[width=\linewidth]{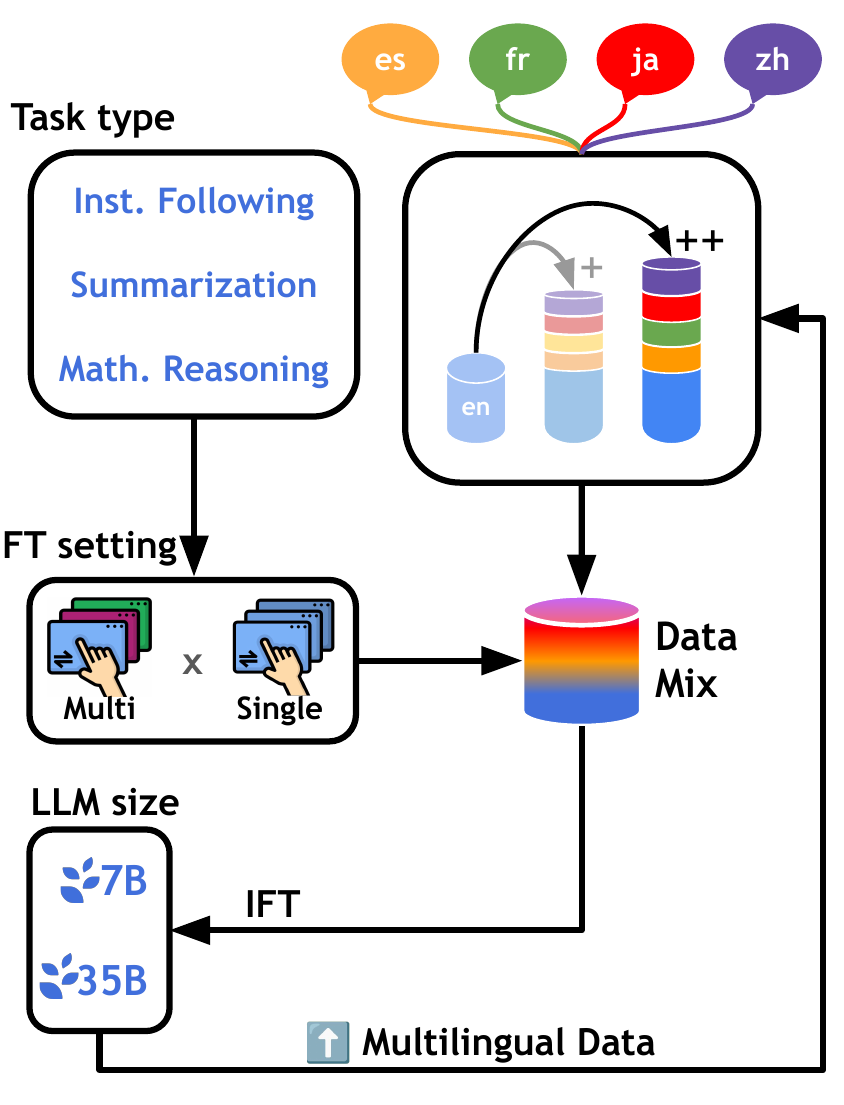}
  \caption{\textbf{Overview of our experimental framework.} We investigate cross-lingual transfer performance improvements during the instruction tuning stage by varying: 1) task type; 2) fine-tuning setting (single-task or multi-task); 3) quantity of multilingual data; and 4) model size. Runs use a fixed amount of English data and increasing amounts of multilingual data.  }
  \label{fig:clt_diagram}
\end{figure}

Instruction tuning is crucial for large language models (LLMs) to develop the ability to follow instructions and to complete more complex tasks. 
\citep{kaplan2020scaling,hoffmann2022training}
Despite its broad adoption and the continued development of strong multilingual LLMs \citep{dubey2024llama,aryabumi2024aya}, performance transfer dynamics during multilingual post-training remain unclear. 
Creating high-quality instruction data is expensive \citep{dubey2024llama}, which makes it infeasible to create large amounts of data in every language for each task. In order to efficiently post-train LLMs in such resource-constrained multilingual settings, they must rely on cross-lingual transfer from the large amount of available English data, combined with limited multilingual data. Understanding the dynamics of cross-lingual transfer during post-training is thus key to building robust models that can excel in increasingly complex tasks across a diverse set of languages.

Prior work has examined cross-lingual transfer (CLT) behavior at an aggregate level \citep{muennighoff-etal-2023-crosslingual,shaham2024multilingual,chen2023monolingual}, but less attention has been given to how dynamics vary across individual tasks, languages, and training scenarios. However, such exploration is critical for developing more robust and generalizable multilingual models, as it allows for targeted improvements that account for the unique challenges posed by specific linguistic and task-related contexts.

In this work, we perform a systematic investigation of multilingual post-training, expanding our current understanding of cross-lingual transfer dynamics.
We examine models' scaling behavior through the lens of multilingual data mixtures during post-training with regard to three important dimensions: task type, model size, and training setting (i.e., single-task or multi-task) across 12 languages. Prior work \citep{kew2023turning,shaham2024multilingual,chen2023monolingual} has been limited by focusing on small-scale models, a single general instruction-following task, and small data budgets. 
We go beyond these by assessing models with both 7B and 35B parameters on 3 tasks with varying complexity levels (summarization, general instruction-following, and mathematical reasoning). Furthermore, we investigate cross-lingual transfer with incrementally increasing multilingual data, using up to 75k unique samples per run, considerably more than prior works. Lastly, we go beyond training on a single task and study the dynamics of training jointly on multiple tasks and languages.

\begin{table*}[]
\centering
\resizebox{\textwidth}{!}{%
\begin{tabular}{llllcc}
\toprule
Task & Train data & Test data & Metric & Seen langs. & Unseen langs. \\ \midrule
Summarization & XLSum & XLSum & RougeL &  &  \\
Instruction Following & ShareGPT & Aya-Dolly200 & LLM-as-a-judge & en,es,fr,ja,zh & ar,ko,pt \\
Mathematical Reasoning & mCoT-Math & MGSM & Accuracy &  &  \\ \bottomrule
\end{tabular}
}
\caption{Dataset references: XLSum \citep{hasan2021xl}, ShareGPT, Aya-Dolly200 \citep{singh2024aya}, mCoT-Math \citep{lai2024mcot}, MGSM \citep{shi2022language}.}
\label{tab:datasets}
\end{table*}

Overall, our findings are the following:
\setlist{nolistsep}
\begin{enumerate}[nolistsep]
\item \textbf{Multilingual performance improvements---but not cross-lingual transfer---are dependent on task type}. Different tasks scale differently when presented with more multilingual data (e.g. mathematical reasoning benefits much more from adding more data) while cross-lingual transfer behaves in a similar way across all tasks. Unseen performance tracks seen performance
and despite CLT 
enabling improvements for the unseen languages it does not enable bridging the performance gap between seen and unseen languages.
\item \textbf{Cross-lingual transfer manifests differently on models of varying scale}. We find that with increasing scale CLT becomes more efficient and the performance gap between seen and unseen languages narrows. At scale, most performance gains are realized using English data only and/or with little multilingual data. 
\item \textbf{Single and multi-task training settings scale differently}. We demonstrate the impact of the training setting where task interference can lead to fluctuations in performance and a wider seen--unseen languages performance gap in multi-task learning. These effects, however, diminish at scale.

\item \textbf{Different tasks benefit from different data mixtures}. While linguistically-oriented tasks (e.g., instruction following, summarization) benefit from data in non-Latin script languages (which tend to have a lower zero-shot performance), for reasoning tasks (e.g., mathematical reasoning) the model learns more efficiently by training on Latin script data only.
\end{enumerate}

\section{Realistic Mulilingual Post-training} \label{sec:experimental-setting}

Our goal is to conduct a comprehensive exploration of multilingual post-training dynamics that reflects the settings of state-of-the-art multilingual LLMs \citep{dubey2024llama,aryabumi2024aya}. We make several assumptions to operationalize these settings in our experiments:

\paragraph{English data availability} We assume that sufficient English data is available for each task. We sample a fixed set of English samples for each task and vary the amount of multilingual data.

\paragraph{Single-task experts vs multi-task generalists} While it is common to adapt models to specific tasks---often via parameter-efficient fine-tuning \citep{pfeiffer2023modular}--- state-of-the-art models are generally trained on many tasks at once \citep{dubey2024llama,aryabumi2024aya}. Consequently, we investigate the cross-lingual transfer dynamics in both single and multi-task instruction tuning.
    
\paragraph{Model scale} Most prior works consider only models up to 7B models. However, as some behavior only emerges at scale \citep{wei2022emergent}, we hypothesize that cross-lingual transfer dynamics also vary with model size. We therefore perform all experiments on both 7B and 35B base models.

\paragraph{Language selection and uniform sampling} Prior work frequently trained jointly on only two languages \citep{de-souza-etal-2024-measuring,zhang2024scaling} while state-of-the-art models are trained on data across multiple languages. In order to ensure comparable results across tasks, we train on English data in addition to a fixed set of 4 languages (Spanish, French, Japanese, and simplified Chinese) from which we sample uniformly.\footnote{Uniform sampling has been shown to be the most efficient sampling strategy in multilingual settings \citep{khanuja-etal-2023-evaluating}.} We evaluate models on these languages as well as a set of unseen languages for each task.

\section{Experimental Setting} \label{sec:experimental_setting}

\begin{figure*}[ht]
    \centering
    \begin{subfigure}{0.48\textwidth}
        \includegraphics[width=\textwidth]{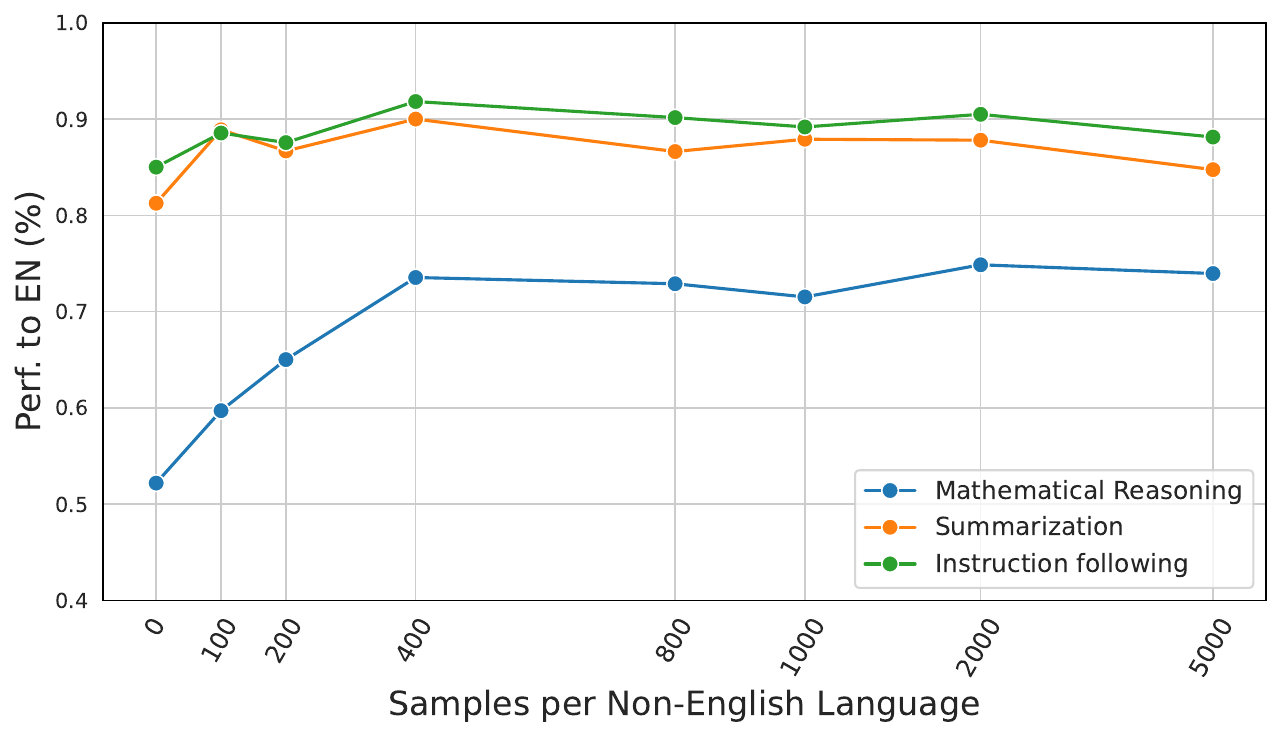}
    \end{subfigure}
    \hfill
    \begin{subfigure}{0.48\textwidth}
        \includegraphics[width=\textwidth]{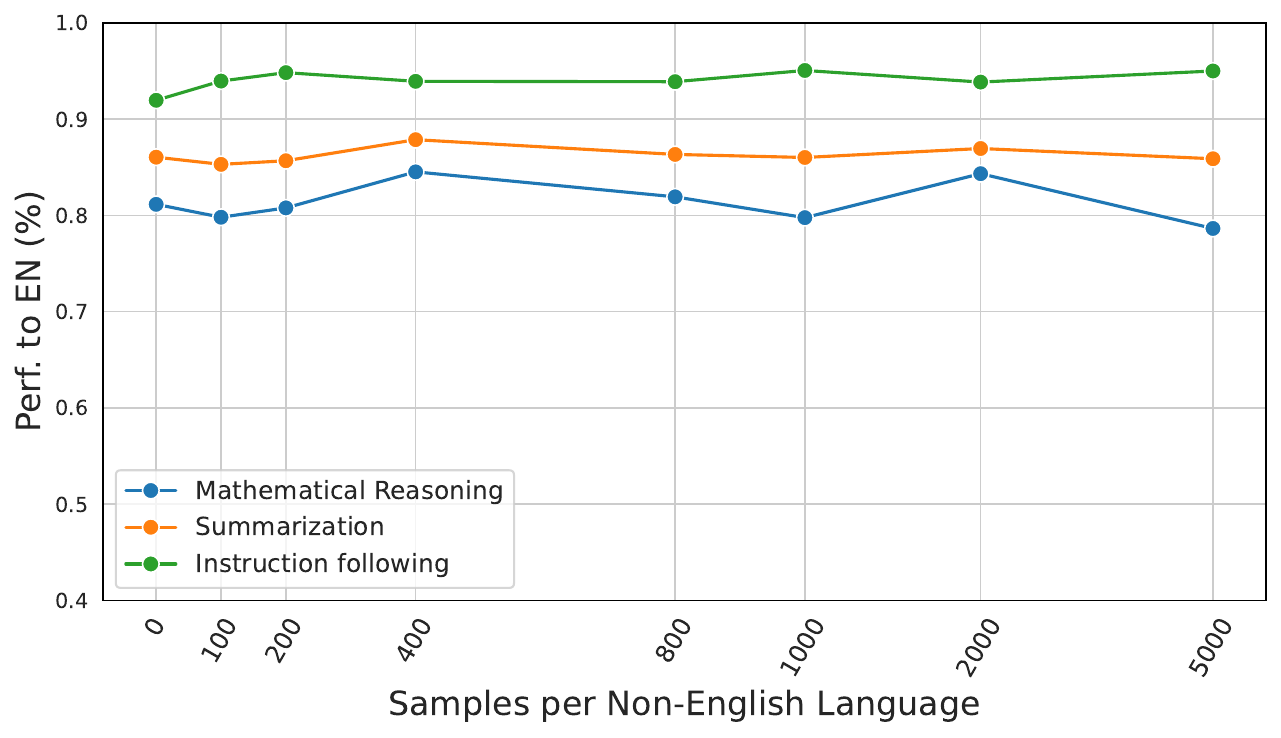}
    \end{subfigure}
    \caption{\textbf{Average performance across seen and unseen languages relative to English across different tasks for 7B (left) and 35B (right) base models} individually (\textbf{single-task}) trained on instruction following (IF), summarization (SM) and mathematical reasoning (MR) datasets. The x-axis indicates the number of samples per non-English language seen during training (es, fr, zh, ja). \textbf{7B (left)}: While IF and summarization results plateau after adding as little as 400 non-English samples per language, MR requires roughly 13x more multilingual data to reach peak performance. \textbf{35B (right)}: We observe similar plateauing behavior as in the smaller model after 200--400 samples across all tasks.}
    \label{fig:relative-to-english-performance}
\end{figure*}

\paragraph{Tasks and Data} We consider three different tasks: summarization, instruction-following, and mathematical reasoning. We select these tasks due to their different complexity levels 
and their availability of training and evaluation data across many languages.
For summarization, we use the XLSum dataset \citep{hasan2021xl} for training and evaluation. For instruction-following, we employ ShareGPT\footnote{\url{https://sharegpt.com/}} prompts with Command-R+\footnote{\url{https://docs.cohere.com/docs/command-r-plus}} completions. 
For mathematical reasoning, we use data from mCoT-math \citep{lai2024mcot} for training and MGSM \citep{shi2022language} for evaluation.\footnote{We translate the MGMS English test set to an additional 3 languages (pt, ko and ar) using Google Translate in order to increase the set of unseen languages available for evaluation.} We provide an overview of all data and languages in Table \ref{tab:datasets}.\footnote{We select the set of seen and unseen languages as the intersection of available training and evaluation languages across tasks respectively.} We conduct experiments on increasing amounts of data for each task, starting with 10k English samples and no non-English samples. With each additional run, we uniformly add samples from the other 4 languages (Spanish, French, Japanese, and simplified Chinese).\footnote{For instance, in the second run, models are trained on 10,400 examples (10k English data + 4$\times$ 100 samples in Spanish, French, Japanese, and Chinese), etc.}\footnote{For the experimental plots, unless stated otherwise, the ``Seen Languages'' category encompasses these four languages and does not include English performance.}

\paragraph{Training} For each experiment, we perform full-model finetuning on the Aya 23 7B and 35B \citep{aryabumi2024aya} and and Llama 3.1 8B base pre-trained checkpoints \citep{dubey2024llama}. 

We determine hyper-parameters based on a grid search. We set a batch size of 64 (unpacked data) and a constant learning rate of 1e-5. We use v5-64 and v5-256 TPUs for 7B/8B and 35B models respectively.
Evaluation checkpoints are saved every 50 steps and selected based on validation loss on a held-out validation set for each run. 

\paragraph{Evaluation} To measure models' instruction-following ability, we generate completions for the 200 human-selected prompts per language of the Dolly200 test set \citep{singh2024aya}. We assign ratings on a scale of 1--5 to each completion using GPT-4o as a pointwise judge following the protocol in \citep{chen-etal-2024-good-data} combined 
with monolingual code-switching metrics \citep{marchisio2024understanding} so as to ensure that models are penalized for answering in a language different from the language of the prompt (see Appendix \ref{appendix:eval_metrics} for details on the ratings). 
For summarization, we compute RougeL \citep{lin-2004-rouge} on 250 randomly selected samples per language from the XLSum \citep{hasan2021xl} test set. Mathematical reasoning performance is measured by calculating the average accuracy of the model on MGSM \citep[8-shot;][]{shi2022language}.

\section{Effects of Task Type and Model Size on Cross-lingual Transfer} \label{sec:task-type_model-size}

We first study the effect of task type and model size on cross-lingual transfer on 7B and 35 models. We fix the amount of unique English samples to 10k and increase the number of instances per non-English training language (French, Spanish, Chinese (simplified), Japanese) from 0 to 5k.
We are particularly interested whether multilingual instruction tuning is able to close the gap to English performance across different tasks. We show the performance averaged across seen and unseen languages relative to English in Figure \ref{fig:relative-to-english-performance} for instruction following, summarization, and mathematical reasoning on Aya 7B and 35B models.

\paragraph{7B results} For the 7B model, we find that some tasks are much more sensitive to the addition of multilingual data than others: while adding non-English instances helps bridge the gap between English and non-English performance by 6.8\% and 8.7\% for instruction following and summarization respectively, mathematical reasoning (MR) performance increases by 22.7\%. We observe similar patterns for base Llama 3.1 8B in Figure \ref{fig:relative2english_llama}, where the English--non-English gap is reduced by 12.3\% for IF and 14.7\% for summarization, while MR displays an improvement of over 40\%. So while multilingual data seems to be particularly effective for MR, small base models also struggle more on multilingual mathematical reasoning relative to English, leaving more room for improvements. 

For mathematical reasoning, the average multilingual performance gradually increases when adding more data  (up to 5k additional instances per non-English language). For the other two tasks, we observe improvements only up to the addition of 400 samples, after which the performance relative to English plateaus. \textbf{Overall, for small models multilingual performance improvements are task-dependent}: on simpler tasks, performance plateaus after 100s of examples per language; on more complex tasks with a larger initial gap to English performance, performance continues to increase and reaches peak performance only after about 13$\times$ more multilingual data (see Tables \ref{tab:10k-english_sharegpt_single_7B}, \ref{tab:10k-english_xlsum_single_7B} and \ref{tab:10k-english_mcot_single_7B}), though returns appear to diminish after the 400 sample mark.

\paragraph{35B results} As can be seen in Figure \ref{fig:relative-to-english-performance}, the addition of multilingual data does not lead to significant improvements for the 35B LLM, as the model trained with English-only data displays very similar performance to models trained with more multilingual data.  \textbf{For large models where the gap to English performance is smaller, multilingual performance plateaus after the addition of 200--400 samples per language for each task.} The difference is particularly stark for mathematical reasoning where the initial gap to English performance is significantly reduced.

\begin{figure}[ht!]
  \centering
  \includegraphics[width=\linewidth]{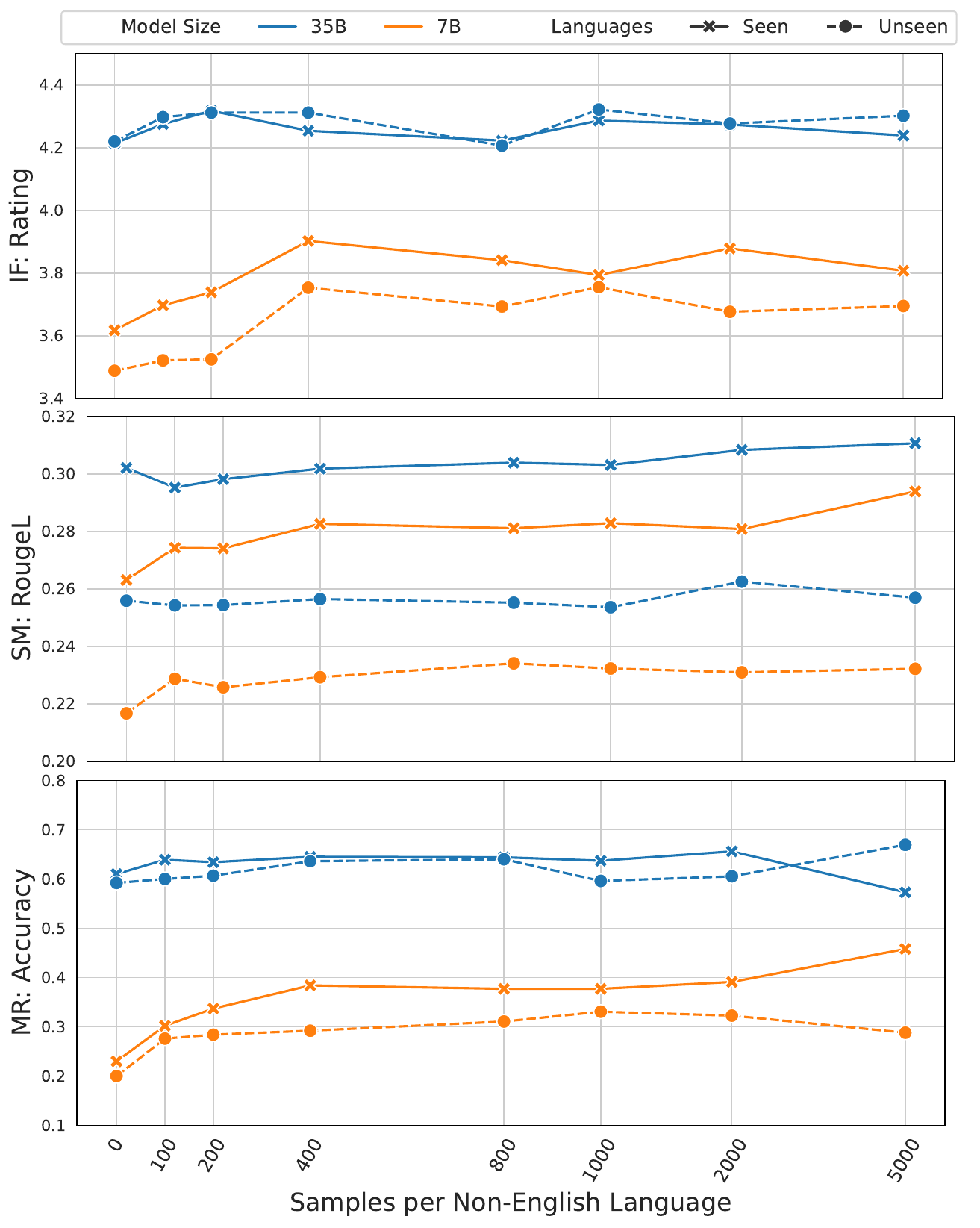}
  \caption{Performance changes across runs with gradually increasing amounts of non-English data for 7B and 35B models \textit{instruction tuned on a single-task setting} on IF, SM and MR tasks, respectively.}
  \label{fig:10k_7B_and_35B_all-tasks}
\end{figure}

\paragraph{Seen vs unseen languages} We have so far reported average performance across seen and unseen languages. However, there is a clear distinction between these two sets, which we illustrate in Figure \ref{fig:10k_7B_and_35B_all-tasks} for 7B and 35B models across the three tasks. As expected, performance on seen languages is generally higher than on unseen languages. However, unseen language performance usually tracks performance on seen languages, especially for smaller models. In other words, \textbf{unseen languages benefit from improvements on seen languages and suffer if seen language performance deteriorates}.

\newcommand\aya{\raisebox{-0.2em}{\includegraphics[width=1em]{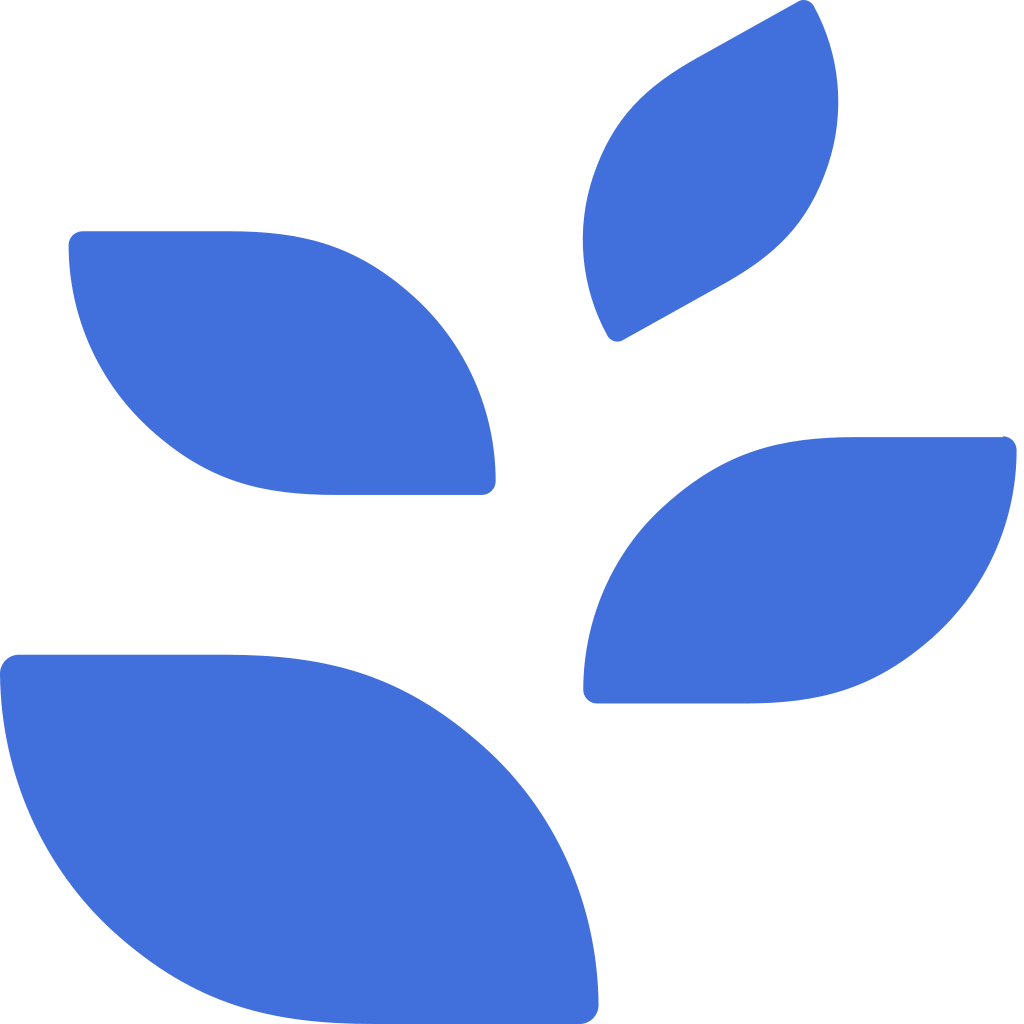}}}

\newcommand\llama{\raisebox{-0.2em}{\includegraphics[width=1em]{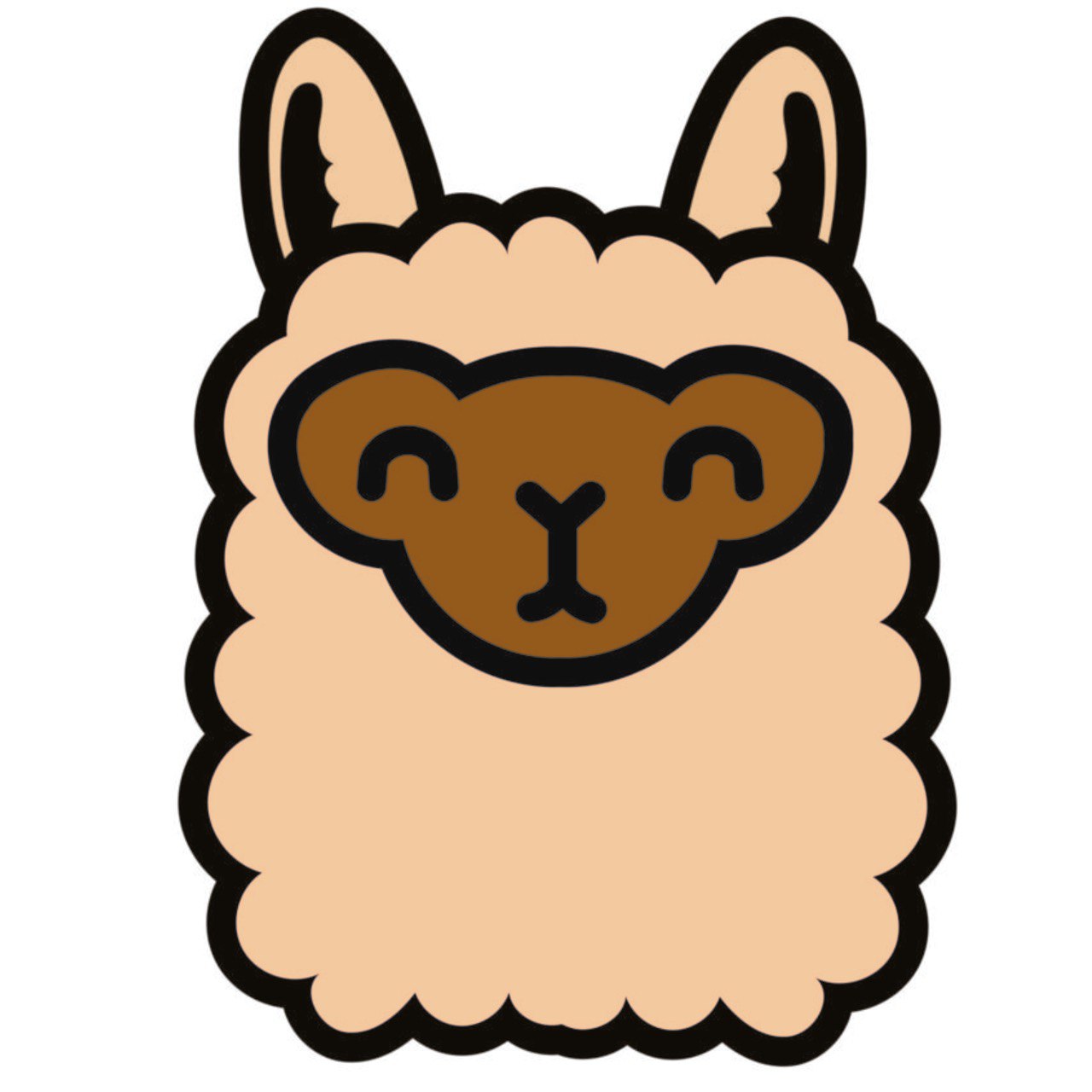}}}

\begin{table}[t]
\centering
\resizebox{0.5\textwidth}{!}{%
\begin{tabular}{llrr}
\toprule
Size & Task & Rank & p-value \\ \midrule
 & Instruction Following & 0.67 & <0.05 \\ 
7B \aya & Summarization & 0.24 & >0.05 \\ 
 & Math. Reasoning & 0.77 & <0.05 \\ \midrule
 & Instruction Following & 0.83 & <0.05 \\ 
8B \llama & Summarization & 0.56 & <0.05 \\ 
 & Math. Reasoning & 0.88 & <0.05 \\ \midrule
 & Instruction Following & 0.51 & <0.05 \\ 
35B \aya & Summarization & 0.39 & <0.05 \\ 
 & Math. Reasoning & 0.49 & <0.05 \\ \bottomrule
\end{tabular}
}
\caption{Spearman's rank correlation coefficient between the average performance of seen and unseen languages during training for Aya \aya$ $ and Llama 3.1 8B \llama$ $ models. \textbf{While there is a significant correlation between seen and unseen performance for smaller models, the correlation decreases with scale}. Furthermore, while there is a strong correlation for the IF and MR tasks, the degree of correlation is weak when it comes to summarization.}
\label{tab:seen_unseen_correlation}
\end{table}

We quantify the correlation between seen and unseen languages performance in Table \ref{tab:seen_unseen_correlation}, which is strong for smaller models and decreases with scale. Furthermore, \textbf{the magnitude of the seen—unseen languages performance gap is model and task-dependent}. For mathematical reasoning and instruction following, there is a performance gap for the small model while the gap is minimal for the large model, indicating improved cross-lingual transfer capabilities. However, for summarization, the performance gap is similarly large for both models.\footnote{While this may indicate challenges in cross-lingual transfer for summarization with large models, it may also point to issues with summarization evaluation \cite{gehrmann2023repairing}.}
Notably, \textbf{across all tasks the initial seen—unseen performance gap after English-only training effectively remains constant}; despite the addition of multilingual data, the model is not able to close the gap to unseen languages. This highlights possible limits of cross-lingual transfer in LLMs.

To account for differences in model architecture we perform the same set of experiments using Llama 3.1 8B \citep{dubey2024llama}, which are represented in Figure \ref{fig:llama_sigle_task}. 
We observe similar results, namely that there is a constant performance gap across all tasks between seen and unseen languages and that the impact of adding more multilingual data depends on the task being analyzed, with mathematical reasoning needing more data to be able to reach peak performance, for instance. One difference however, is that while behavior remains similar the magnitude of gains obtained when adding multilingual data is larger for the Llama 3.1 8B base models. We hypothesize this is due to the model having been pre-trained on a smaller amount of multilingual data. It therefore obtains a lower multilingual average performance when instruction-tuned on English samples only and requires more multilingual data to fully leverage its capabilities across its supported languages.

\section{Single vs Multi-task Instruction Tuning} \label{single-vs-multi}

After investigating cross-lingual transfer behavior in a single-task setting, we now take a look at the dynamics in a multi-task setting. The latter is both more challenging---as not only languages but also tasks may interfere with each other---and more realistic, as LLMs are usually trained on a variety of tasks at a time. 

For these experiments we consider the three tasks analyzed in the previous section (i.e. instruction following, summarization and mathematical reasoning). We fix the number of unique English samples to 15k (5k from each task) and gradually increase the number of instances per non-English language uniformly across tasks from 0 (English only data) up to 7k.

\begin{figure*}[ht]
    \centering
    \begin{subfigure}{0.48\textwidth}
        \includegraphics[width=\textwidth]{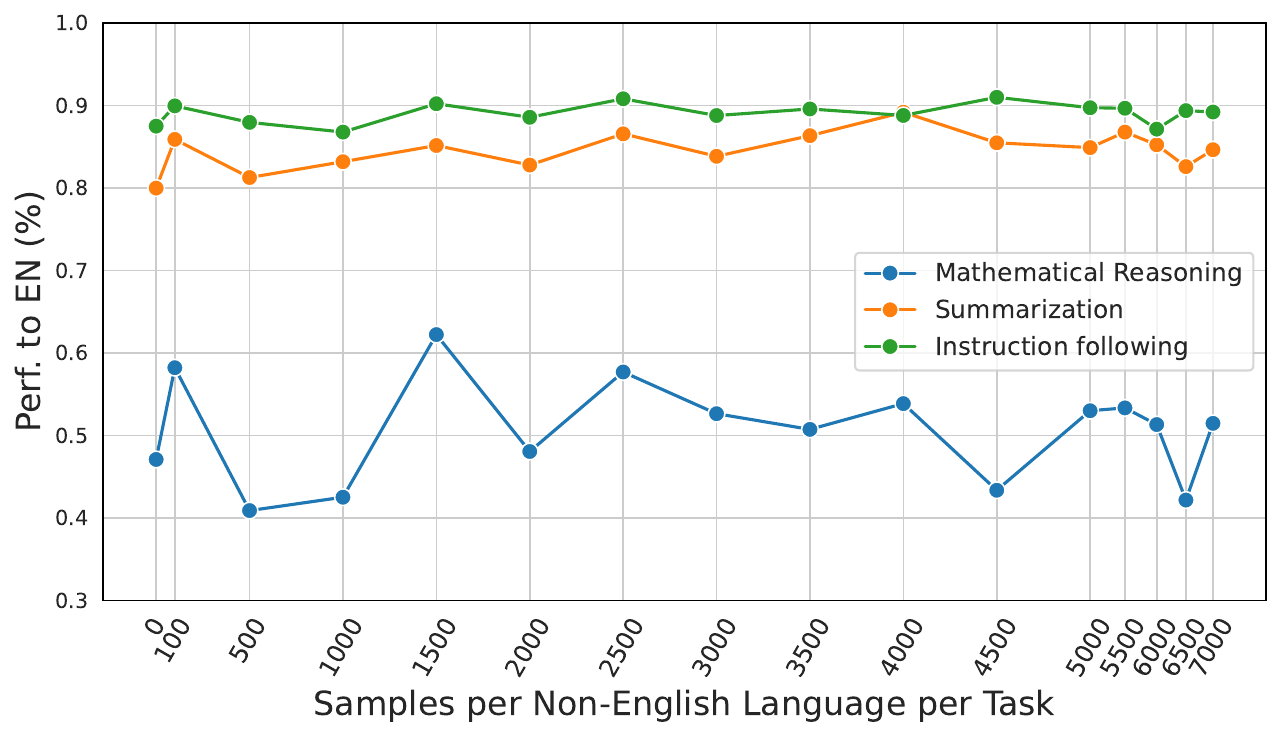}
    \end{subfigure}
    \hfill
    \begin{subfigure}{0.48\textwidth}
        \includegraphics[width=\textwidth]{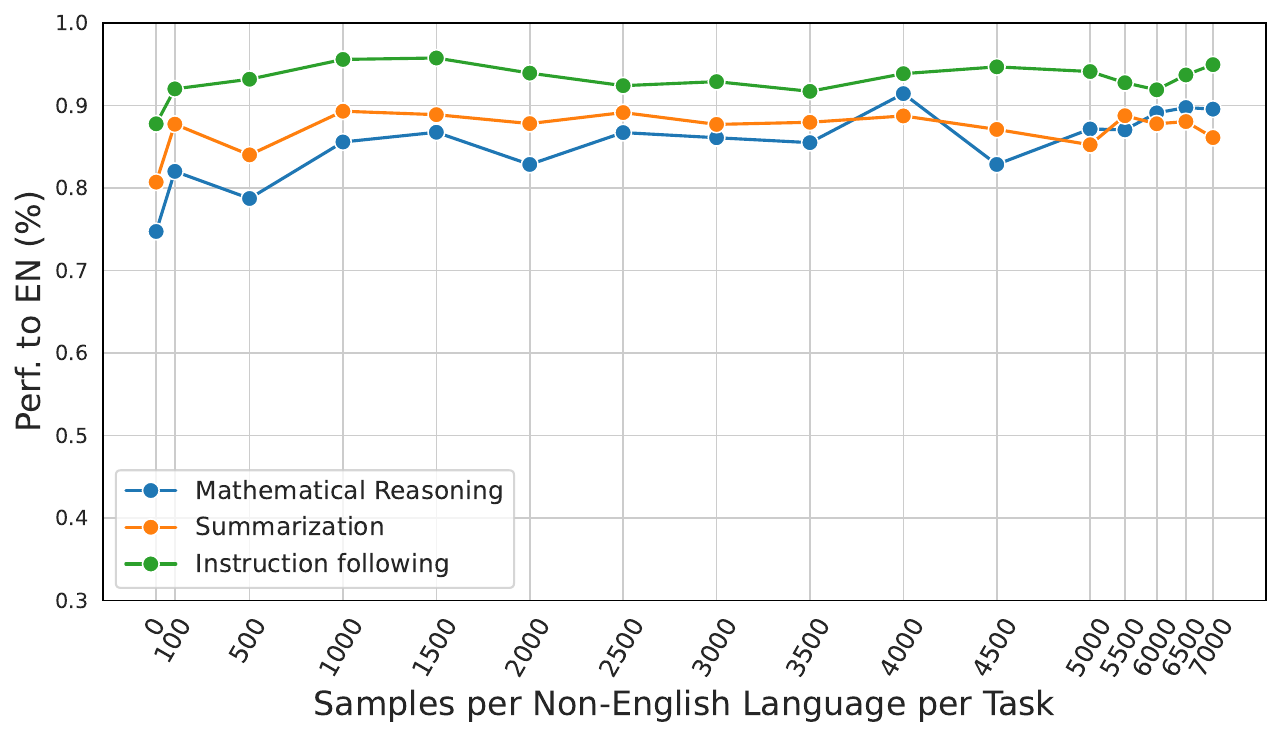}
    \end{subfigure}
    \caption{\textbf{Average performance across seen and unseen languages relative to English across different tasks for 7B (left) and 35B (right) base models} trained jointly (\textbf{multi-task}) on instruction following (IF), summarization (SM) and mathematical reasoning (MR) datasets. The x-axis indicates the number of samples per non-English language seen during training (es, fr, zh, ja). \textbf{7B (left)}: While IF and summarization results manage to reach an average of 90\% performance relative to English, MR displays a hectic improvement behavior, reaching only a little over 60\% relative performance, which represents a decrease of over 10\% when compared to single task results in Figure \ref{fig:relative-to-english-performance}. \textbf{35B (right)}: We observe similar plateauing behavior as in the smaller model after 200--400 samples across all tasks, with a very similar performance relative to English for all tasks which seems to narrow even further in the multi-task setting when compard to numbers in Figure \ref{fig:relative-to-english-performance}.}
    \label{fig:relative-to-english-performance-mixed}
\end{figure*}

We show results relative to English for Aya 7B and 35B models trained jointly on the three tasks in Figure \ref{fig:relative-to-english-performance-mixed}. Comparing those changes to the ones in Figure \ref{fig:relative-to-english-performance}, we can make two main observations for the 7B model: first, performance oscillates more when training jointly on multiple tasks. Secondly, while non-reasoning tasks do not suffer performance drops, mathematical reasoning is negatively affected by the other two tasks.

\begin{figure}[ht!]
  \centering
  \includegraphics[width=\linewidth]{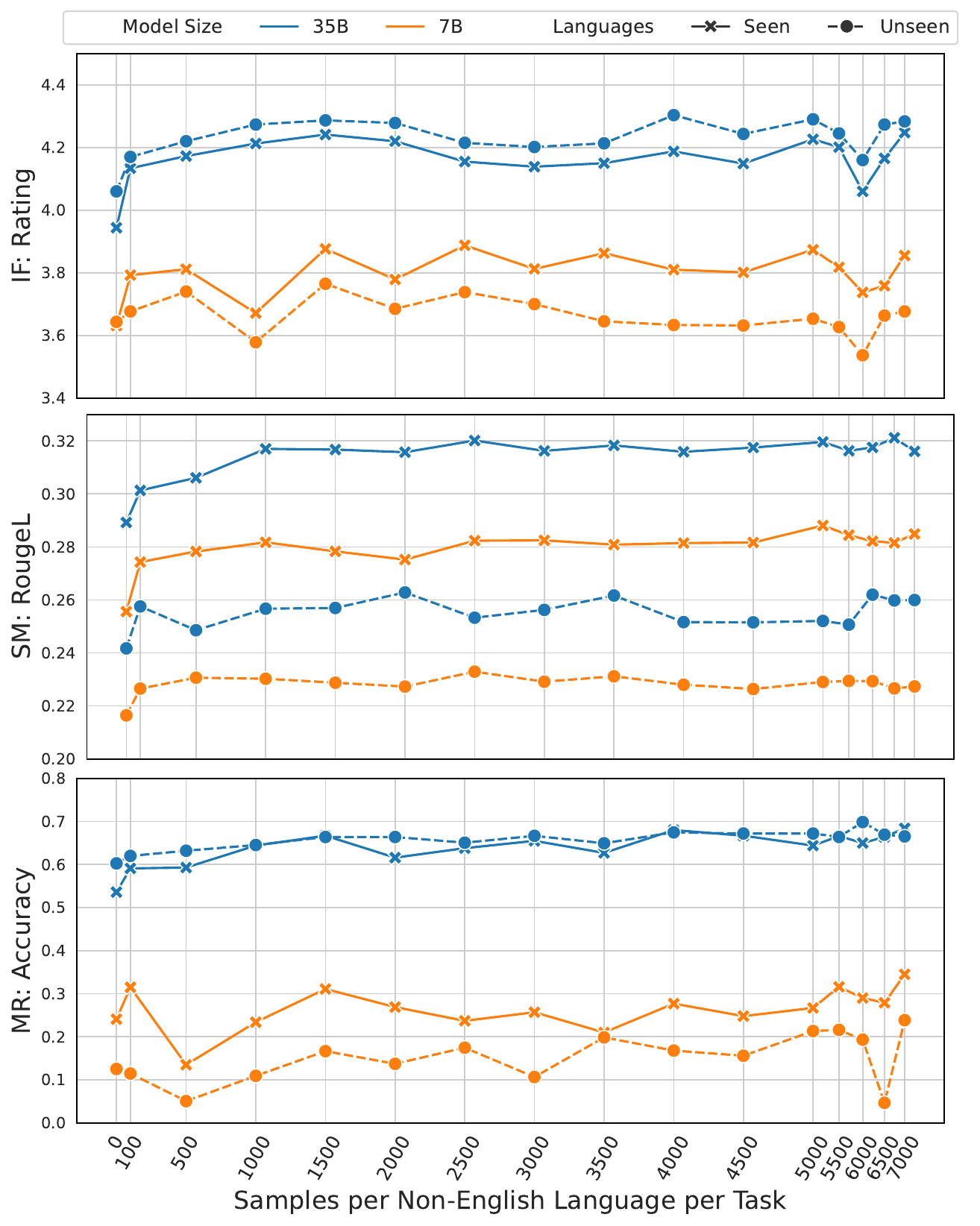}
  \caption{Performance changes across runs with gradually increasing amounts of non-English data for 7B and 35B models \textit{instruction tuned on a multi-task setting} on IF, SM and MR tasks, respectively.} 
  \label{fig:10k_7B_and_35B_all-tasks-mixed}
\end{figure}

\subsection{Multi-task learning at scale} \label{sec:multitask_at_scale}

We can observe the benefits of scale if we compare the results of the 7B with the 35B model in Figure \ref{fig:relative-to-english-performance-mixed}. The 35B model achieves bigger gains across non-English languages with an increasing proportion of multilingual data, whereas the downstream performance of its smaller counterpart across the three tasks plateaus quickly. Additionally, we can see that while there is considerable oscillation in terms of mathematical reasoning performance on MGSM for the 7B model, the 35B LLM does not suffer from the same issue as its math performance gradually improves throughout the runs.

The three plots in Figure \ref{fig:10k_7B_and_35B_all-tasks-mixed} compare the absolute performance of multi-task 7B and 35B models considering instruction following, summarization and mathematical reasoning, respectively, from top to bottom. For IF (uppermost plot), we observe that the seen--unseen gap increases with the addition of more multilingual data for the 7B model as displayed in Figure \ref{fig:seen_unseen_gap}. For summarization (SM; middle plot), while 7B performance remains the same both in the single and multi-task settings, for the larger LLM we observe the model outperforms the best run in the single task setting on seen languages 
despite seeing half of the amount of task-specific multilingual data (2.5k samples per task per language in the multi-task setting as compared to 5k per language in the single-task one), indicating that SM benefits from this mixed training setting at scale.

For mathematical reasoning (Figure \ref{fig:10k_7B_and_35B_all-tasks-mixed}, bottom), we can observe that the larger model does not seem to suffer negative interference from the data from other tasks and gradually improves with the introduction of more multilingual data. Conversely, the math performance of the 7B model oscillates considerably, with a slight improvement over time.

Additionally, we observe that while the 7B model degrades in MGSM performance by 8\% (considering best average performance across all runs) when trained on a multi-task data mix, the 35B LLM benefits from the task diversity, matching the peak accuracy obtained in the single-task setting (68\%) with less task-specific data (11k total mCoT-math samples in the multi-task setting compared to 18k samples in the single-task one). Overall, \textbf{non-English languages benefit from task diversity in multi-task learning in large models.} 

Finally, while in a single-task setting (see Figure \ref{fig:10k_7B_and_35B_all-tasks}), the average downstream performance of seen languages is usually larger than that of unseen ones, in the mixed task setting the opposite happens for the IF and MR tasks with the 35B model. That is, the performance on unseen languages becomes slightly higher than that on seen languages for most runs. This is promising as it indicates that \textbf{in realistic multi-task conditions with sufficient scale, models are able to effectively transfer to unseen languages for some tasks.}

\section{Varying Language Script Proportions}

So far, we kept the set of training languages fixed and only varied their proportions in the data. We now study the impact of this selection and ask: \textit{How does grouping languages into different training mixtures affect cross-lingual transfer?} 

We consider two different training mixtures: non-Latin script languages (en, zh, ja) and Latin-script languages (en, es, fr). Separating languages based on scripts enables the creation of two significantly distinct mixtures, allowing us to explore both \textbf{(1)} the impact of grouping languages with the same scripts (Latin script mixture); and \textbf{(2)} the impact of prioritizing languages with lower general performance that the model might have seen less during pre-training (e.g., zh, ja).

We use the same experimental setting where we incrementally add non-English data to the instruction tuning mix (from 0 to 5k samples per non-English language) while keeping the number of English samples fixed at 10k instances. 
We report and analyze the average performance over two categories (excluding English): Latin script languages (es, fr, pt) and non-Latin script languages (ja, zh, ko, ar). 

\begin{figure}[t]
    \centering
    \begin{subfigure}[t]{0.48\textwidth}
        \centering
        \includegraphics[width=1.\textwidth]{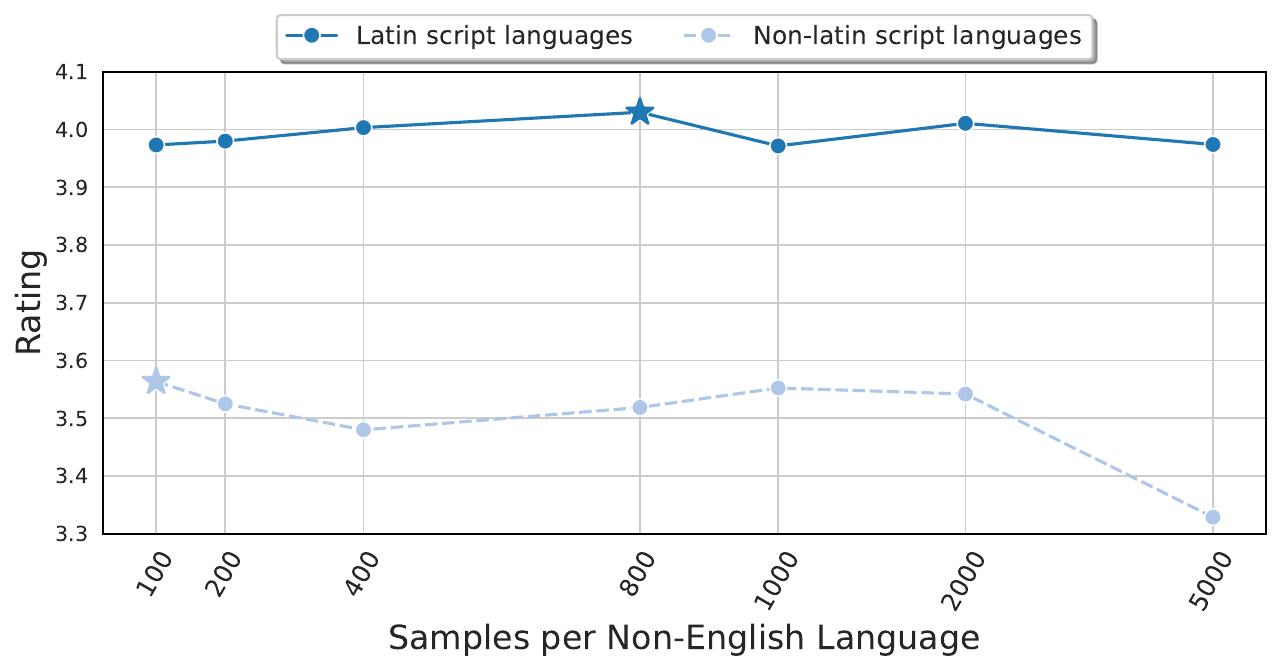}
    \end{subfigure}%
    
    \begin{subfigure}[t]{0.48\textwidth}
        \centering
        \includegraphics[width=1.\textwidth]{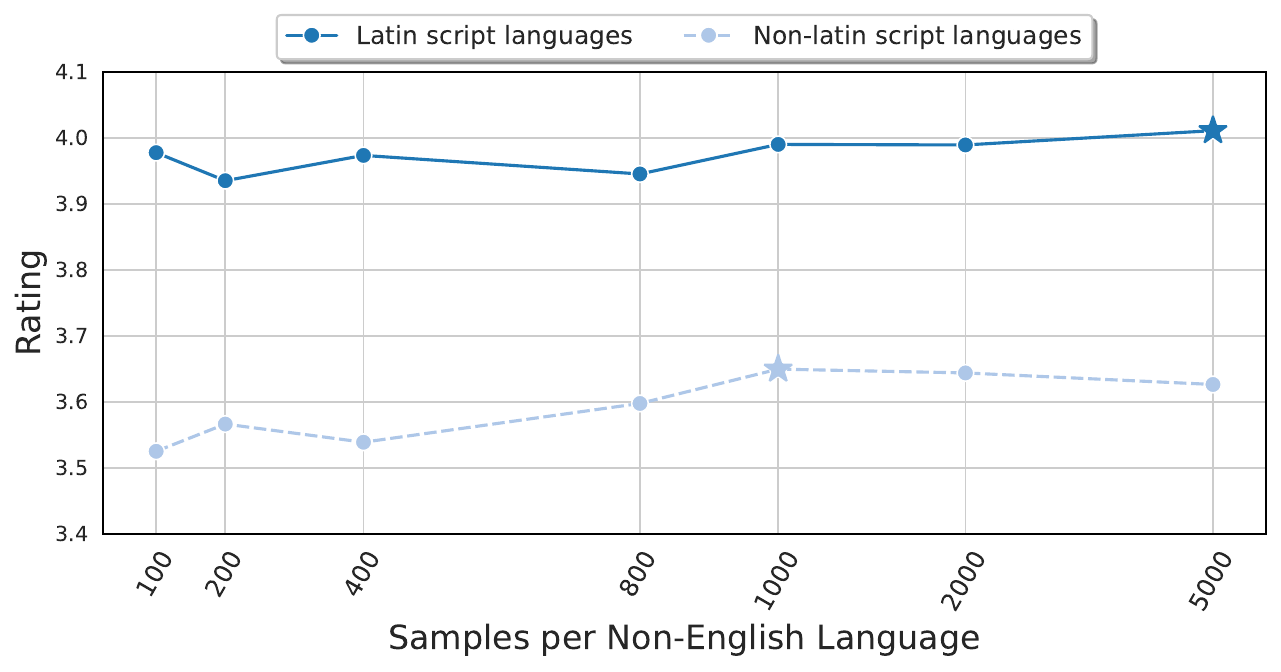}
    \end{subfigure}
    
  \caption{Instruction following performance of Aya 7B base models instruction tuned on a data mix comprised of a \textbf{[en + es, fr] mix (top)} and another trained on a \textbf{[en + ja, zh] mix (bottom)}.}
  \label{fig:script-sharegpt}
\end{figure}

\begin{figure}[ht!]
    \centering
    \begin{subfigure}[t]{0.48\textwidth}
        \centering
        \includegraphics[width=1.\textwidth]{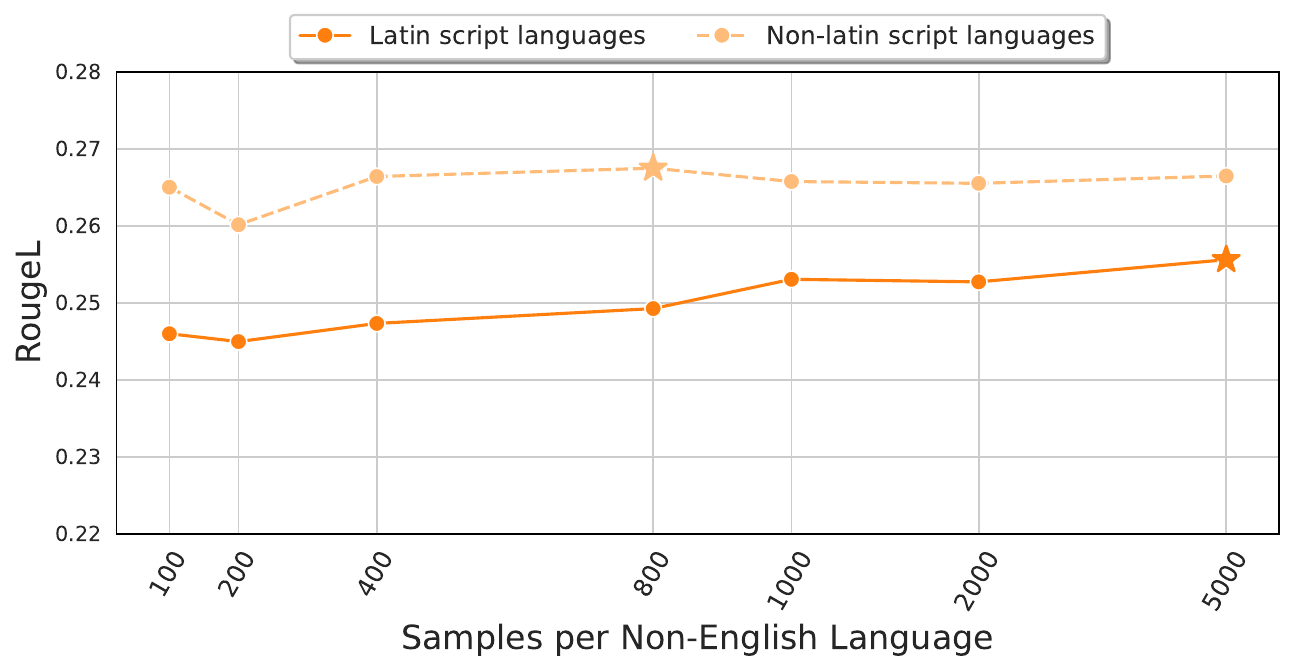}
    \end{subfigure}%
    
    \begin{subfigure}[t]{0.48\textwidth}
        \centering
        \includegraphics[width=1.\textwidth]{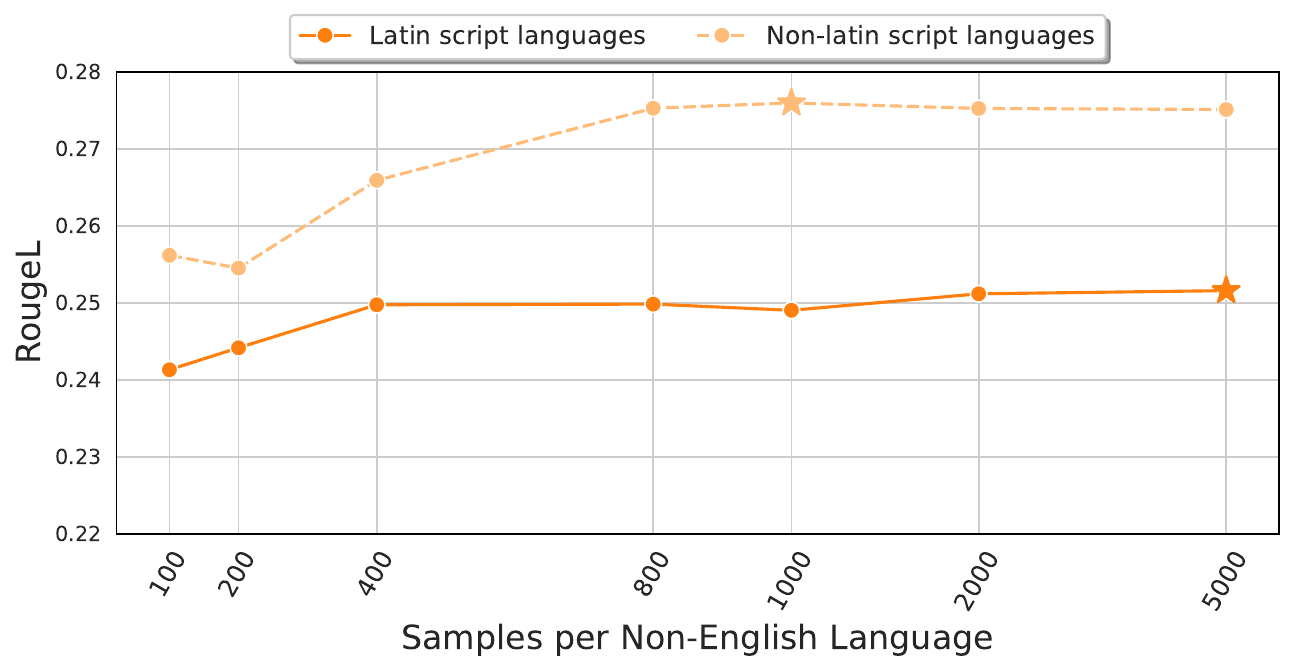}
    \end{subfigure}
    
  \caption{Summarization performance of Aya 7B base models instruction tuned on a data mix comprised of a \textbf{[en + es, fr] mix (top)} and another trained on a \textbf{[en + ja, zh] mix (bottom)}.}
  \label{fig:script-xlsum}
\end{figure}

Figure \ref{fig:script-sharegpt} and \ref{fig:script-xlsum} show the performance progression for instruction following and summarization, respectively. In both cases, performance on Latin script languages (solid lines) follows a similar pattern of improvement across runs regardless of the language mix used. On the other hand, training on non-Latin script languages seems to be necessary to leverage performance for that group of languages, that is, \textbf{cross-lingual transfer from Latin script languages seems to be insufficient to significantly improve performance across languages that make use of other writing systems.}

\begin{figure}[t]
    \centering
    \begin{subfigure}[t]{0.48\textwidth}
        \centering
        \includegraphics[width=1.\textwidth]{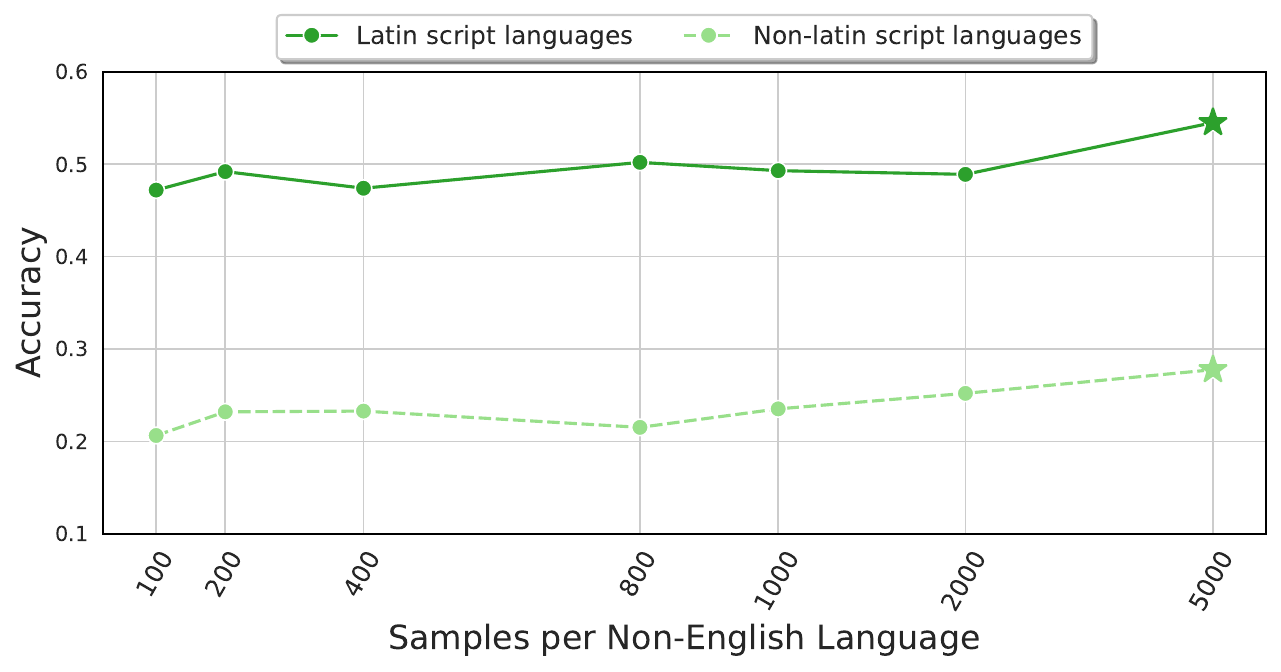}
    \end{subfigure}%
    
    \begin{subfigure}[t]{0.48\textwidth}
        \centering
        \includegraphics[width=1.\textwidth]{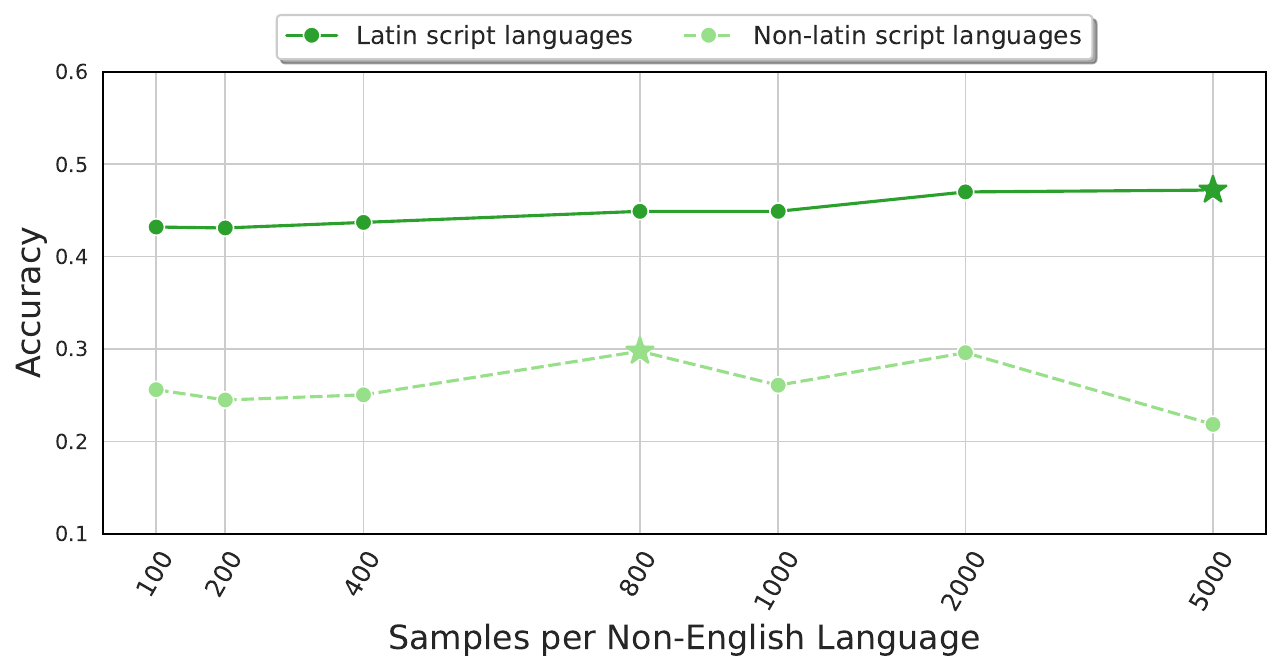}
    \end{subfigure}
    
  \caption{Mathematical reasoning performance of Aya 7B base models instruction tuned on a data mix comprised of a \textbf{[en + es, fr] mix (top)} and another trained on a \textbf{[en + ja, zh] mix (bottom)}.}
  \label{fig:script-mcot}
\end{figure}

However, we observe different trends for the mathematical reasoning task displayed in Figure \ref{fig:script-mcot}. Comparing the plots, we can see that not only does non-Latin script performance track closely Latin script performance, but reasoning capabilities seem to be better learned when using Latin script data only (36.6\% vs 37.4\%, when comparing peak average performance across all languages),
approaching the average performance of the original experiments (38.4\%; see Figure \ref{fig:10k_7B_and_35B_all-tasks}) despite using 10k less samples (as in this setting we consider 2 non-English languages compared to 4 in the original setting). This might indicate that \textbf{while linguistically-oriented tasks (e.g., IF and SM) require data from non-Latin script languages to leverage performance for that linguistic group, reasoning tasks (e.g., MR) may benefit from  Latin-script data mixes during post-training, relying less on the presence of other languages}.

\section{Related Work}

With instruction tuning emerging as a standard post-training step \citep{weller-etal-2020-learning, wei2021finetuned, mishra2021cross, sanh2021multitask}, applications of instruction tuning to the multilingual setting started to receive attention.
Initial approaches showed the benefits of cross-lingual transfer for both seen and unseen languages \citep{muennighoff-etal-2023-crosslingual, kew2023turning}. Subsequent work focused on analyzing different areas including comparing monolingual vs multilingual instruction tuning \citep{chen2023monolingual,shaham2024multilingual}, finding that instruction tuning with even fairly small data across multiple languages is effective, even for English-centric models \citep{kew2023turning,ye2023language}. 
Multilingual instruction tuning has also been found beneficial for mitigating adverse behavior such as language confusion \citep{marchisio2024understanding}. Other studies focused on understanding the impact of translated vs human-annotated data and the impact of translation artefacts \citep{kew2023turning,chen-etal-2024-good-data}. 
Finally, some studies seek to establish scaling laws in LLM fine-tuning, focusing on controlled, bilingual settings \citep{de-souza-etal-2024-measuring,zhang2024scaling}.
While the above studies provide valuable insights into the dynamics of multilingual instruction tuning and cross-lingual transfer, they are considerably constrained in terms of data and computational budgets, mostly experimenting with quantized small-scale LLMs (of up to 8B parameters) while mainly focusing on open-ended chat. In addition, they frequently employ parameter-efficient fine-tuning methods such as LoRA \citep{hu2021lora, dettmers2024qlora} instead of full-parameter finetuning. This paper goes beyond prior work by considering large models and data budgets, tasks with varying complexity, and realistic training conditions reflecting both single and multi-task instruction tuning.

\section{Conclusion}

In this work, we perform a systematic and comprehensive study to provide better insights into the workings of cross-lingual transfer and multilingual learning overall across many languages and post-training settings. We show that the benefits of multilingual data are different across tasks of varying type and complexity. Moreover, we find that scaling up models can be effective in reducing the performance gap between seen and unseen languages and that larger models leverage CLT more efficiently. Lastly, we show that the dynamics of cross-lingual transfer become much more unstable during multi-task learning and different individual tasks seem to benefit more from distinct language mixes.  Overall, our findings contribute to a deeper understanding of CLT and multilingual instruction tuning and we hope it can inspire future work on this topic.

\section{Limitations}

Our work offers valuable insights into the dynamics of cross-lingual transfer and multilingual instruction tuning. However, it is important to acknowledge several limitations that may affect the generalizability of our findings, and we highlight these as opportunities for future research.

A key limitation lies in the scope of languages included in our experimental setup. To ensure reliable comparisons of results and trends across tasks, we restricted the number of unseen languages during training. This restriction was further compounded by the limited availability of test sets for certain languages. Additionally, the tasks we considered were constrained by the availability of training and evaluation data across the selected set of languages: English (en), Spanish (es), French (fr), Japanese (ja), Chinese (zh), Arabic (ar), Korean (ko), and Portuguese (pt).

Another limitation stems from computational constraints, which led us to limit the number of experimental runs and the variety of models evaluated. While these decisions were driven by practical considerations, they may impact the robustness and generalizability of our findings.

\bibliography{custom}

\appendix

\section{Evaluation Metrics} \label{appendix:eval_metrics}

\paragraph{Switch-aware ratings} The quality and helpfulness of responses is measured by using automatic ratings, where GPT-4o is used as a judge and asked to give a rating ranging from 1 to 5 based on the criteria defined in the prompt in Figure \ref{fig:rating_prompt}. Furthermore, since the responses are being evaluated in a multilingual setting using monolingual prompts (i.e., question and answer are expected to be in the same language) we assign responses that are in the wrong language a score of 1, by calculating the binary pass rate for each completion \citep{marchisio2024understanding}. This is done for the purpose of performing more comprehensive evaluation of the generations, as LLMs with poor multilingual capabilities tend to answer in English regardless of the language used in the instruction, while a judge model may still assign a high score to English responses given its English-centric bias \citep{marchisio2024understanding}. 

\begin{figure}[ht!]
\centering
\begin{tikzpicture}[scale=1.0, every node/.style={transform shape}]
\node[rectangle, rounded corners, draw=blue!20, fill=blue!10, text width=\linewidth, align=left, inner sep=1.2ex, font=\ttfamily] (prompt) {
Please act as an impartial judge and evaluate the quality of the response provided by an AI assistant to the user instruction displayed 
below. Your evaluation should consider factors such as helpfulness, relevance, accuracy, depth, creativity, and level of detail. It is also required that the response is in the same language as the instruction. 
\linebreak
\linebreak
Begin your evaluation with a short explanation. Be as objective as possible. After providing your explanation, please rate the response on a scale of 1 to 5 by strictly following this format:“[[rating]]”, for example: “Rating: [[2]]”
\linebreak

[Question]
\linebreak
\{question\}
\linebreak

[The Start of Assistant's Answer]
\linebreak
\{generation\}

[The End of Assistant's Answer]

};

\end{tikzpicture}
\caption{
Prompt used to automatically evaluate models' responses to the Dolly200 prompts.
}
\label{fig:rating_prompt}
\end{figure}

\newpage

\section{Llama 3.1 8B Results} \label{appendix:add_results}

Results for experiments using using Llama 3.1 8B model.

\begin{figure}[ht!]
  \centering
  \includegraphics[width=\linewidth]{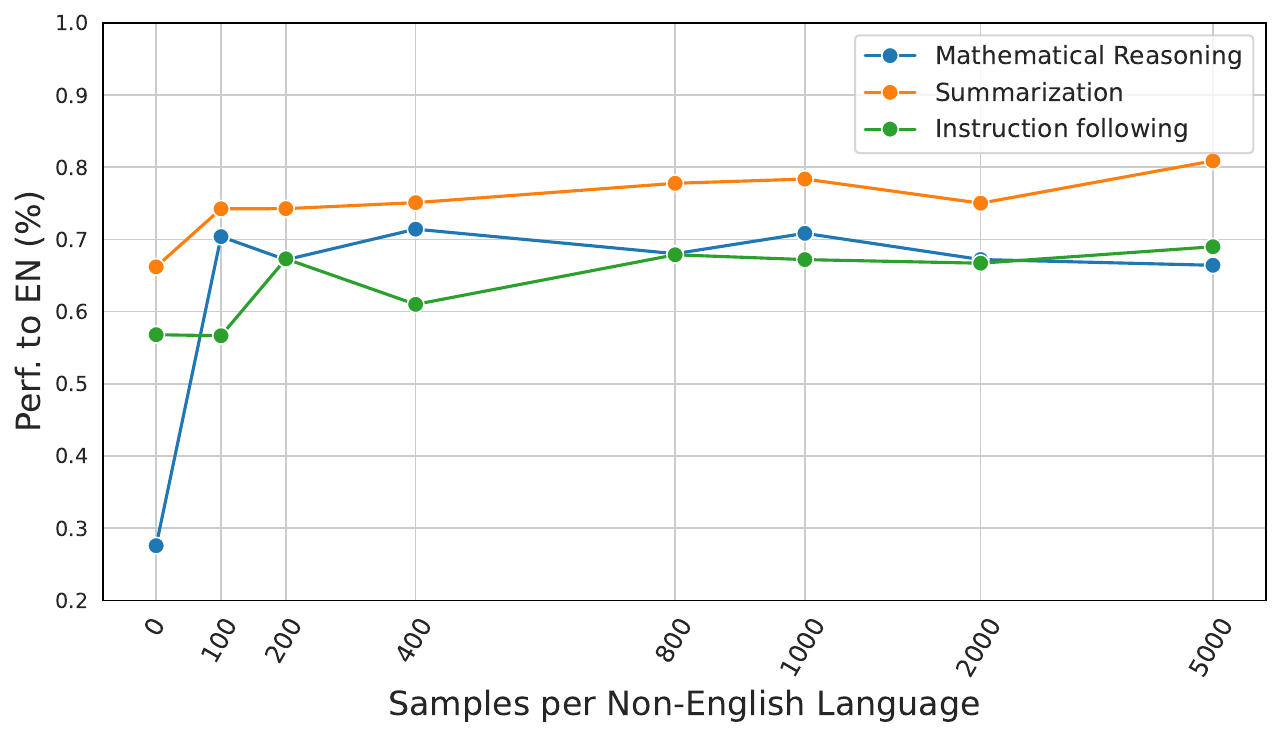}
  \caption{\textbf{Average performance across seen and unseen languages relative to English across different tasks for base Llama 3.1 8B} individually (\textbf{single-task}) trained on instruction following (IF), summarization (SM) and mathematical reasoning (MR) datasets. The x-axis indicates the number of samples per non-English language seen during training (es, fr, zh, ja).}
  \label{fig:relative2english_llama}
\end{figure}

\begin{figure}[ht!]
    \centering
    \begin{subfigure}[t]{0.48\textwidth}
        \centering
        \includegraphics[width=1.\textwidth]{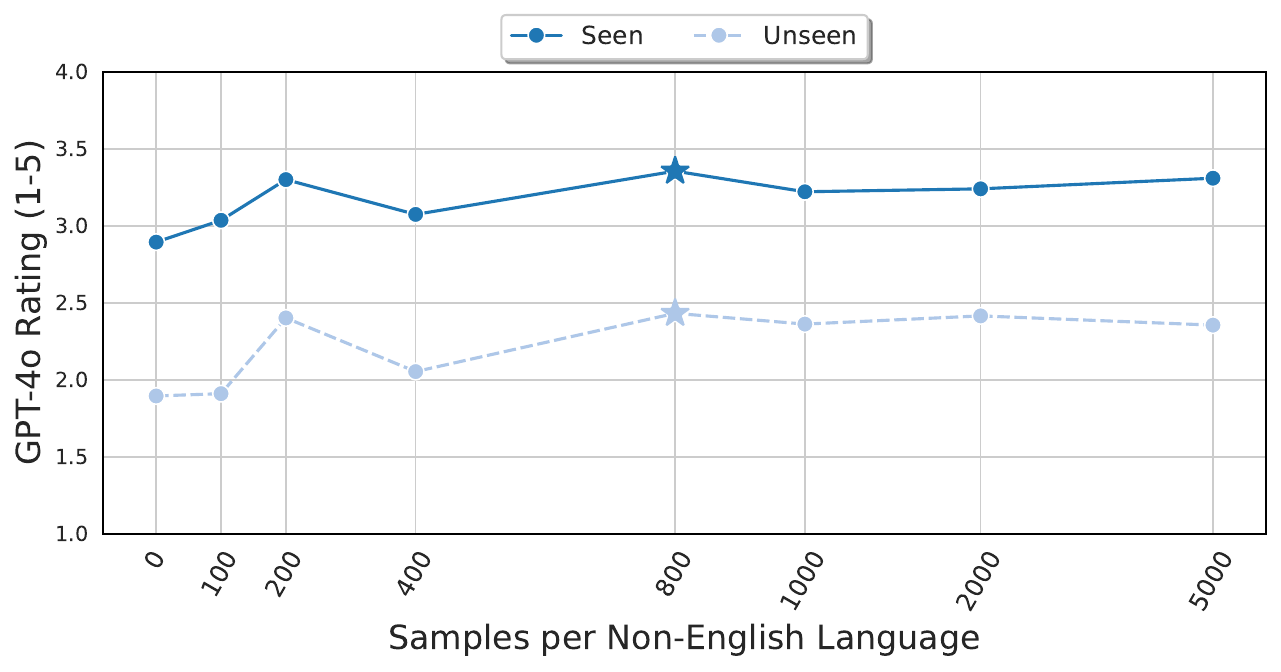}
    \end{subfigure}%
    
    \begin{subfigure}[t]{0.48\textwidth}
        \centering
        \includegraphics[width=1.\textwidth]{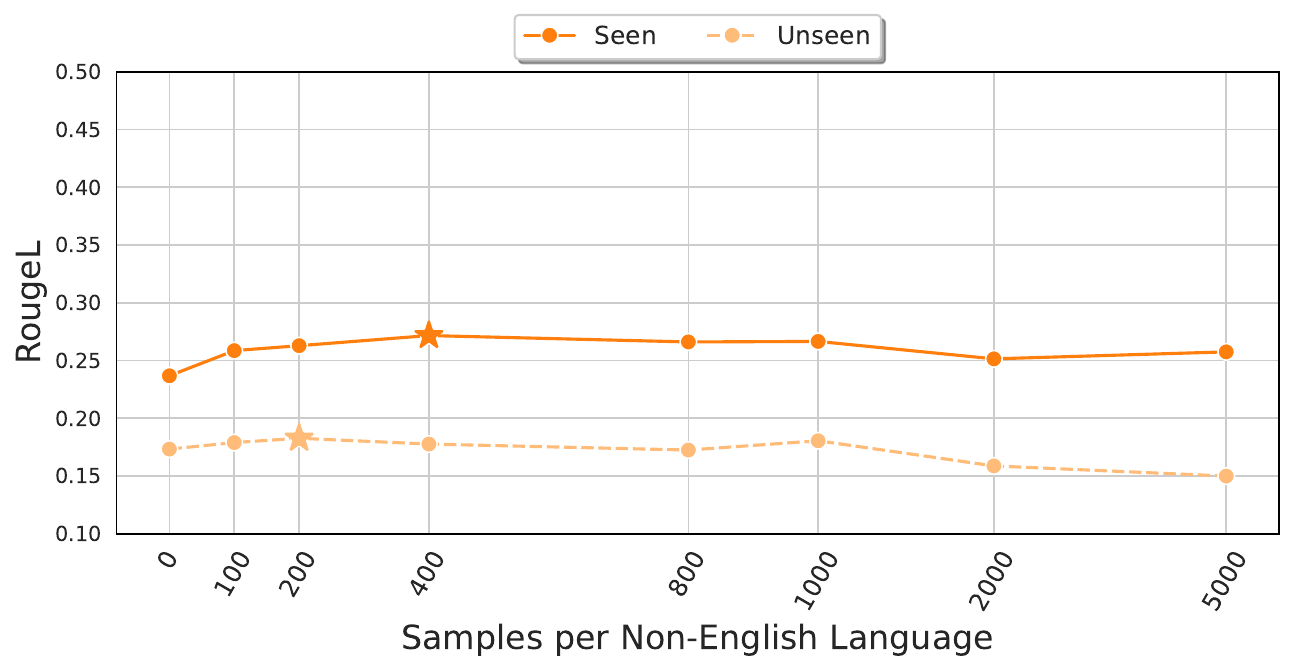}
    \end{subfigure}

    \begin{subfigure}[t]{0.48\textwidth}
        \centering
        \includegraphics[width=1.\textwidth]{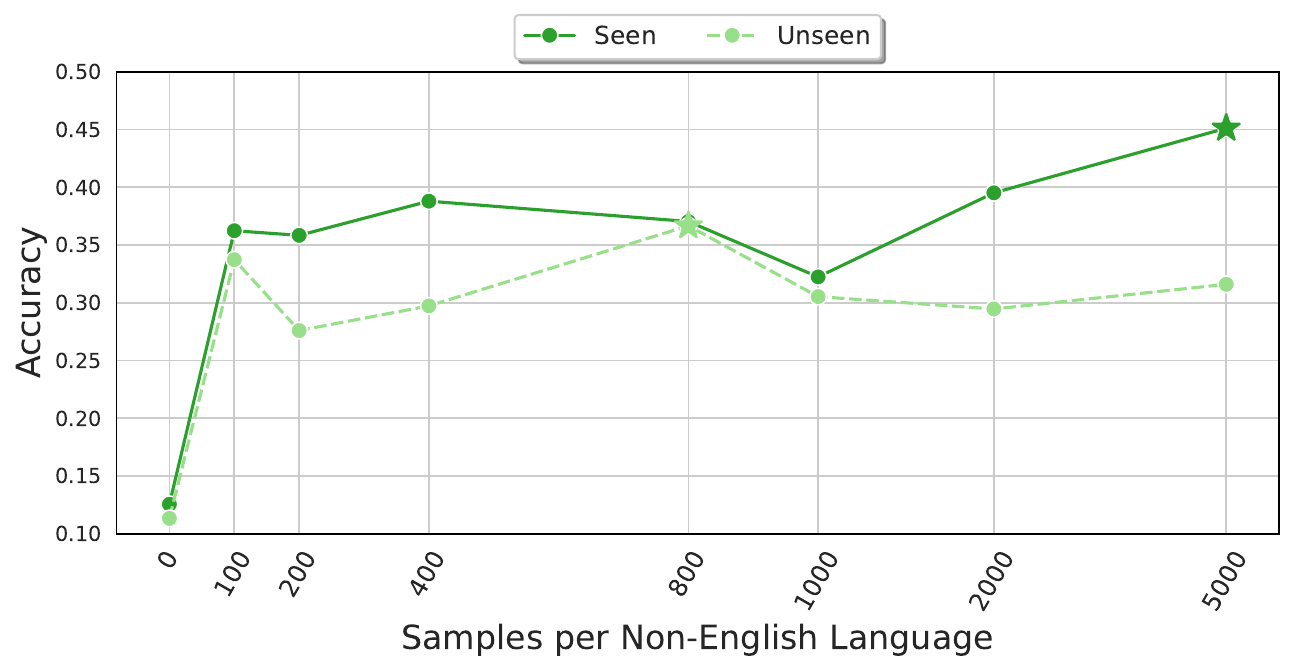}
    \end{subfigure}
  \caption{Performance changes across runs with gradually increasing amounts of non-English data for base Llama 3.1 8B \textit{instruction tuned on a single-task setting} on IF, SM and MR tasks, respectively.}
  \label{fig:llama_sigle_task}
\end{figure}

\section{Downstream Performance Per Language Per Task} \label{appendix:add_results}

Tables \ref{tab:10k-english_sharegpt_single_35B} through \ref{tab:10k-english_mcot_multi_7B} display the average downstream performance per language per task for all experiments discussed in the main paper. The values correspond to the metrics presented in Section \ref{sec:experimental_setting}. 

\subsection{Instruction Following}

Tables \ref{tab:10k-english_sharegpt_single_35B} through \ref{tab:10k-english_sharegpt_multi_7B} present the average automatic ratings, calculated per language, for different finetuned model responses to the prompts of the Dolly200 test set \citep{singh2024aya}.

\begin{table}[ht!]
\centering
\resizebox{0.5\textwidth}{!}{%
\begin{tabular}{l cccccccc}
\toprule
 & \textbf{0} & \textbf{100} & \textbf{200} & \textbf{400} & \textbf{800} & \textbf{1000} & \textbf{2000} & \textbf{5000} \\
\midrule
\textbf{ar} & 4.20 & 4.25 & 4.26 & 4.28 & 4.17 & 4.35 & 4.28 & 4.30 \\
\textbf{en} & 4.58 & 4.56 & 4.55 & 4.55 & 4.49 & 4.53 & 4.55 & 4.49 \\
\textbf{es} & 4.44 & 4.43 & 4.54 & 4.42 & 4.41 & 4.48 & 4.51 & 4.41 \\
\textbf{fr} & 4.48 & 4.54 & 4.57 & 4.48 & 4.45 & 4.53 & 4.53 & 4.48 \\
\textbf{ja} & 4.16 & 4.17 & 4.18 & 4.08 & 4.04 & 4.25 & 4.03 & 4.11 \\
\textbf{ko} & 4.13 & 4.30 & 4.37 & 4.38 & 4.25 & 4.39 & 4.24 & 4.36 \\
\textbf{pt} & 4.33 & 4.34 & 4.31 & 4.28 & 4.20 & 4.22 & 4.32 & 4.24 \\
\textbf{zh} & 3.77 & 3.95 & 3.98 & 4.03 & 3.99 & 3.89 & 4.03 & 3.96 \\
\midrule
\textbf{Avg.} & 4.26 & 4.32 & 4.34 & 4.31 & 4.25 & 4.33 & 4.31 & 4.29 \\
\bottomrule
\end{tabular}}
\caption{Automatic switch-aware ratings (1-5) for base 35B Aya model instruction tuned on a single-task setting on [en, es, fr, ja, zh]. The columns represent the numbers of samples per non-English language used to finetune the model.}
\label{tab:10k-english_sharegpt_single_35B}
\end{table}

\begin{table}[ht!]
\centering
\resizebox{0.5\textwidth}{!}{%
\begin{tabular}{l cccccccc}
\toprule
 & \textbf{0} & \textbf{100} & \textbf{200} & \textbf{400} & \textbf{800} & \textbf{1000} & \textbf{2000} & \textbf{5000} \\
\midrule
\textbf{ar} & 3.31 & 3.31 & 3.38 & 3.63 & 3.58 & 3.62 & 3.50 & 3.43 \\
\textbf{en} & 4.19 & 4.09 & 4.17 & 4.18 & 4.19 & 4.24 & 4.19 & 4.26 \\
\textbf{es} & 3.90 & 3.94 & 3.92 & 4.03 & 4.04 & 4.03 & 4.03 & 4.08 \\
\textbf{fr} & 3.99 & 4.05 & 4.20 & 4.11 & 4.14 & 4.13 & 4.26 & 4.18 \\
\textbf{ja} & 3.48 & 3.39 & 3.44 & 3.85 & 3.68 & 3.61 & 3.60 & 3.73 \\
\textbf{ko} & 3.44 & 3.58 & 3.44 & 3.73 & 3.58 & 3.77 & 3.67 & 3.81 \\
\textbf{pt} & 3.71 & 3.67 & 3.75 & 3.90 & 3.92 & 3.88 & 3.85 & 3.85 \\
\textbf{zh} & 3.10 & 3.40 & 3.39 & 3.62 & 3.50 & 3.40 & 3.63 & 3.24 \\
\midrule
\textbf{Avg.} & 3.64 & 3.68 & 3.71 & 3.88 & 3.83 & 3.83 & 3.84 & 3.82 \\
\bottomrule
\end{tabular}}
\caption{Automatic switch-aware ratings (1-5) for base 7B Aya model instruction tuned on a single-task setting on [en, es, fr, ja, zh]. The columns represent the numbers of samples per non-English language used to finetune the model.}
\label{tab:10k-english_sharegpt_single_7B}
\end{table}

\begin{table}[ht!]
\centering
\resizebox{0.5\textwidth}{!}{%
\begin{tabular}{l cccccccc}
\toprule
 & \textbf{0} & \textbf{100} & \textbf{200} & \textbf{400} & \textbf{800} & \textbf{1000} & \textbf{2000} & \textbf{5000} \\
\midrule
\textbf{ar} & 3.45 & 3.54 & 3.56 & 3.54 & 3.50 & 3.59 & 3.56 & 3.24 \\
\textbf{en} & 4.20 & 4.14 & 4.17 & 4.20 & 4.25 & 4.17 & 4.17 & 4.18 \\
\textbf{es} & 3.98 & 3.98 & 3.99 & 3.98 & 4.12 & 4.01 & 4.09 & 4.01 \\
\textbf{fr} & 4.01 & 4.08 & 4.11 & 4.20 & 4.20 & 4.18 & 4.21 & 4.20 \\
\textbf{ja} & 3.46 & 3.52 & 3.44 & 3.36 & 3.50 & 3.42 & 3.67 & 3.37 \\
\textbf{ko} & 3.69 & 3.71 & 3.63 & 3.67 & 3.63 & 3.71 & 3.67 & 3.66 \\
\textbf{pt} & 3.83 & 3.85 & 3.87 & 3.84 & 3.87 & 3.81 & 3.90 & 3.85 \\
\textbf{zh} & 3.30 & 3.49 & 3.46 & 3.36 & 3.45 & 3.48 & 3.25 & 3.04 \\
\midrule
\textbf{Avg.} & 3.74 & 3.79 & 3.78 & 3.77 & 3.81 & 3.80 & 3.82 & 3.69 \\
\bottomrule
\end{tabular}}
\caption{Automatic switch-aware ratings (1-5) for base Aya 7B model instruction tuned on a single-task setting on [en, es, fr]. The columns represent the numbers of samples per non-English language used to finetune the model.}
\label{tab:latin-script_10k-english_sharegpt_script-categories}
\end{table}

\begin{table}[ht!]
\centering
\resizebox{0.5\textwidth}{!}{%
\begin{tabular}{l cccccccc}
\toprule
 & \textbf{0} & \textbf{100} & \textbf{200} & \textbf{400} & \textbf{800} & \textbf{1000} & \textbf{2000} & \textbf{5000} \\
\midrule
\textbf{ar} & 3.33 & 3.42 & 3.48 & 3.54 & 3.57 & 3.63 & 3.60 & 3.62 \\
\textbf{en} & 4.25 & 4.20 & 4.20 & 4.24 & 4.22 & 4.14 & 4.21 & 4.18 \\
\textbf{es} & 3.99 & 4.10 & 3.94 & 4.06 & 3.98 & 4.01 & 3.96 & 3.98 \\
\textbf{fr} & 4.08 & 4.04 & 4.04 & 4.10 & 4.09 & 4.12 & 4.18 & 4.20 \\
\textbf{ja} & 3.52 & 3.67 & 3.67 & 3.56 & 3.68 & 3.69 & 3.73 & 3.62 \\
\textbf{ko} & 3.52 & 3.65 & 3.56 & 3.55 & 3.63 & 3.77 & 3.67 & 3.75 \\
\textbf{pt} & 3.77 & 3.90 & 3.83 & 3.69 & 3.81 & 3.90 & 3.83 & 3.88 \\
\textbf{zh} & 3.24 & 3.35 & 3.55 & 3.50 & 3.50 & 3.50 & 3.58 & 3.52 \\
\midrule
\textbf{Avg.} & 3.71 & 3.79 & 3.79 & 3.78 & 3.81 & 3.85 & 3.85 & 3.84 \\
\bottomrule
\end{tabular}}
\caption{Automatic switch-aware ratings (1-5) for base Aya 7B model instruction tuned on a single-task setting on [en, ja, zh]. The columns represent the numbers of samples per non-English language used to finetune the model.}
\label{tab:non-latin-script_10k-english_sharegpt_script}
\end{table}

\begin{table}[ht!]
\centering
\resizebox{0.5\textwidth}{!}{%
\begin{tabular}{l cccccccc}
\toprule
 & \textbf{0} & \textbf{100} & \textbf{200} & \textbf{400} & \textbf{800} & \textbf{1000} & \textbf{2000} & \textbf{5000} \\
\midrule
\textbf{ar} & 1.44 & 1.42 & 1.92 & 1.42 & 1.81 & 1.83 & 1.86 & 1.71 \\
\textbf{en} & 4.05 & 4.21 & 4.15 & 4.08 & 4.18 & 4.07 & 4.13 & 4.05 \\
\textbf{es} & 3.38 & 3.35 & 3.71 & 3.40 & 3.69 & 3.64 & 3.62 & 3.71 \\
\textbf{fr} & 3.13 & 3.67 & 3.71 & 3.56 & 3.75 & 3.64 & 3.79 & 3.83 \\
\textbf{ja} & 1.71 & 1.76 & 2.28 & 2.00 & 2.41 & 2.15 & 2.11 & 2.35 \\
\textbf{ko} & 1.54 & 1.50 & 2.08 & 1.71 & 2.23 & 2.02 & 1.99 & 1.98 \\
\textbf{pt} & 2.70 & 2.81 & 3.21 & 3.04 & 3.26 & 3.24 & 3.39 & 3.38 \\
\textbf{zh} & 2.21 & 2.19 & 2.65 & 2.33 & 2.75 & 2.62 & 2.54 & 2.60 \\
\midrule
\textbf{Avg.} & 2.52 & 2.61 & 2.96 & 2.69 & 3.01 & 2.90 & 2.93 & 2.95 \\
\bottomrule
\end{tabular}}
\caption{Automatic switch-aware ratings (1-5) for base Llama 3.1 8B model instruction tuned on a single-task setting on [en, es, fr, ja, zh]. The columns represent the numbers of samples per non-English language used to finetune the model.}
\label{tab:10k-english_sharegpt_llama31-8B_table}
\end{table}

\begin{table*}[ht!]
\centering
\resizebox{1.\textwidth}{!}{%
\begin{tabular}{l cccccccccccccccc}
\toprule
 & \textbf{0} & \textbf{100} & \textbf{500} & \textbf{1000} & \textbf{1500} & \textbf{2000} & \textbf{2500} & \textbf{3000} & \textbf{3500} & \textbf{4000} & \textbf{4500} & \textbf{5000} & \textbf{5500} & \textbf{6000} & \textbf{6500} & \textbf{7000} \\
\midrule
\textbf{ar} & 4.00 & 4.20 & 4.22 & 4.28 & 4.30 & 4.30 & 4.30 & 4.28 & 4.29 & 4.31 & 4.25 & 4.29 & 4.25 & 4.20 & 4.33 & 4.36 \\
\textbf{en} & 4.55 & 4.51 & 4.50 & 4.43 & 4.45 & 4.52 & 4.53 & 4.49 & 4.55 & 4.51 & 4.42 & 4.52 & 4.55 & 4.46 & 4.50 & 4.49 \\
\textbf{es} & 4.39 & 4.43 & 4.45 & 4.45 & 4.50 & 4.45 & 4.51 & 4.39 & 4.45 & 4.46 & 4.43 & 4.46 & 4.50 & 4.44 & 4.46 & 4.43 \\
\textbf{fr} & 4.45 & 4.42 & 4.51 & 4.49 & 4.50 & 4.47 & 4.56 & 4.50 & 4.49 & 4.51 & 4.47 & 4.50 & 4.46 & 4.43 & 4.50 & 4.53 \\
\textbf{ja} & 3.75 & 4.03 & 4.06 & 4.11 & 4.09 & 4.11 & 4.08 & 4.06 & 4.08 & 4.08 & 4.09 & 4.20 & 4.10 & 3.98 & 4.04 & 4.18 \\
\textbf{ko} & 3.94 & 4.04 & 4.15 & 4.21 & 4.26 & 4.26 & 4.04 & 4.07 & 4.10 & 4.34 & 4.21 & 4.32 & 4.21 & 4.01 & 4.24 & 4.25 \\
\textbf{pt} & 4.24 & 4.28 & 4.29 & 4.33 & 4.29 & 4.27 & 4.29 & 4.26 & 4.25 & 4.25 & 4.26 & 4.26 & 4.26 & 4.27 & 4.26 & 4.25 \\
\textbf{zh} & 3.19 & 3.65 & 3.67 & 3.81 & 3.88 & 3.85 & 3.47 & 3.60 & 3.59 & 3.69 & 3.60 & 3.75 & 3.75 & 3.38 & 3.67 & 3.85 \\
\midrule
\textbf{Avg.} & 4.06 & 4.19 & 4.23 & 4.26 & 4.28 & 4.28 & 4.22 & 4.21 & 4.22 & 4.27 & 4.22 & 4.29 & 4.26 & 4.15 & 4.25 & 4.29 \\
\bottomrule
\end{tabular}}
\caption{Automatic switch-aware ratings (1-5) for base Aya 35B model instruction tuned on a multi-task setting on [en, es, fr, ja, zh]. The columns represent the numbers of samples per non-English language from ShareGPT used to finetune the model.}
\label{tab:10k-english_sharegpt_multi_35B}
\end{table*}

\begin{table*}[ht!]
\centering
\resizebox{1.\textwidth}{!}{%
\begin{tabular}{l cccccccccccccccc}
\toprule
 & \textbf{0} & \textbf{100} & \textbf{500} & \textbf{1000} & \textbf{1500} & \textbf{2000} & \textbf{2500} & \textbf{3000} & \textbf{3500} & \textbf{4000} & \textbf{4500} & \textbf{5000} & \textbf{5500} & \textbf{6000} & \textbf{6500} & \textbf{7000} \\
\midrule
\textbf{ar} & 3.44 & 3.44 & 3.60 & 3.41 & 3.67 & 3.60 & 3.64 & 3.57 & 3.48 & 3.52 & 3.37 & 3.44 & 3.52 & 3.31 & 3.40 & 3.50 \\
\textbf{en} & 4.20 & 4.17 & 4.30 & 4.21 & 4.27 & 4.24 & 4.21 & 4.25 & 4.24 & 4.22 & 4.11 & 4.22 & 4.20 & 4.21 & 4.18 & 4.26 \\
\textbf{es} & 4.08 & 3.98 & 4.01 & 3.94 & 4.17 & 4.04 & 4.12 & 3.99 & 4.12 & 4.06 & 3.99 & 4.04 & 4.08 & 4.09 & 4.04 & 4.11 \\
\textbf{fr} & 3.98 & 4.00 & 4.12 & 4.04 & 4.12 & 4.09 & 4.14 & 4.30 & 4.20 & 4.21 & 4.13 & 4.14 & 4.13 & 4.12 & 4.14 & 4.11 \\
\textbf{ja} & 3.38 & 3.75 & 3.62 & 3.36 & 3.74 & 3.63 & 3.70 & 3.62 & 3.79 & 3.65 & 3.65 & 3.77 & 3.62 & 3.60 & 3.57 & 3.69 \\
\textbf{ko} & 3.62 & 3.79 & 3.81 & 3.56 & 3.81 & 3.71 & 3.73 & 3.65 & 3.66 & 3.60 & 3.64 & 3.69 & 3.60 & 3.53 & 3.67 & 3.73 \\
\textbf{pt} & 3.87 & 3.81 & 3.81 & 3.76 & 3.82 & 3.75 & 3.84 & 3.88 & 3.79 & 3.78 & 3.88 & 3.83 & 3.75 & 3.77 & 3.91 & 3.81 \\
\textbf{zh} & 3.09 & 3.42 & 3.49 & 3.35 & 3.48 & 3.35 & 3.58 & 3.33 & 3.35 & 3.33 & 3.44 & 3.54 & 3.44 & 3.14 & 3.27 & 3.52 \\
\midrule
\textbf{Avg.} & 3.71 & 3.80 & 3.85 & 3.70 & 3.88 & 3.80 & 3.87 & 3.83 & 3.83 & 3.80 & 3.78 & 3.83 & 3.79 & 3.72 & 3.78 & 3.84 \\
\bottomrule
\end{tabular}}
\caption{Automatic switch-aware ratings (1-5) for base Aya 7B model instruction tuned on a multi-task setting on [en, es, fr, ja, zh]. The columns represent the numbers of samples per non-English language from ShareGPT used to finetune the model.}
\label{tab:10k-english_sharegpt_multi_7B}
\end{table*}
\FloatBarrier

\subsection{Summarization}

Tables \ref{tab:10k-english_xlsum_single_35B} through \ref{tab:10k-english_xlsum_multi_7B} present the average RougeL score, calculated per language, for summaries generated by different finetuned models using the XLSum test set \citep{hasan2021xl}.

\begin{table}[ht!]
\centering
\resizebox{0.5\textwidth}{!}{%
\begin{tabular}{l cccccccc}
\toprule
 & \textbf{0} & \textbf{100} & \textbf{200} & \textbf{400} & \textbf{800} & \textbf{1000} & \textbf{2000} & \textbf{5000} \\
\midrule
\textbf{ar} & 0.27 & 0.26 & 0.27 & 0.26 & 0.26 & 0.27 & 0.28 & 0.28 \\
\textbf{en} & 0.33 & 0.33 & 0.33 & 0.32 & 0.33 & 0.33 & 0.33 & 0.33 \\
\textbf{es} & 0.24 & 0.20 & 0.20 & 0.21 & 0.21 & 0.21 & 0.22 & 0.21 \\
\textbf{fr} & 0.30 & 0.28 & 0.28 & 0.29 & 0.30 & 0.31 & 0.29 & 0.31 \\
\textbf{ja} & 0.38 & 0.40 & 0.40 & 0.40 & 0.40 & 0.39 & 0.41 & 0.42 \\
\textbf{ko} & 0.20 & 0.20 & 0.20 & 0.21 & 0.20 & 0.21 & 0.20 & 0.20 \\
\textbf{pt} & 0.29 & 0.30 & 0.30 & 0.30 & 0.30 & 0.29 & 0.31 & 0.29 \\
\textbf{zh} & 0.29 & 0.30 & 0.31 & 0.31 & 0.30 & 0.31 & 0.31 & 0.31 \\
\midrule
\textbf{Avg.} & 0.29 & 0.28 & 0.29 & 0.29 & 0.29 & 0.29 & 0.29 & 0.29 \\
\bottomrule
\end{tabular}}
\caption{RougeL score calculated for base Aya 35B model instruction tuned on a single-task setting on [en, es, fr, ja, zh]. The columns represent the numbers of samples per non-English language used to finetune the model.}
\label{tab:10k-english_xlsum_single_35B}
\end{table}

\begin{table}[ht!]
\centering
\resizebox{0.5\textwidth}{!}{%
\begin{tabular}{l cccccccc}
\toprule
 & \textbf{0} & \textbf{100} & \textbf{200} & \textbf{400} & \textbf{800} & \textbf{1000} & \textbf{2000} & \textbf{5000} \\
\midrule
\textbf{ar} & 0.24 & 0.25 & 0.24 & 0.25 & 0.26 & 0.26 & 0.25 & 0.26 \\
\textbf{en} & 0.30 & 0.29 & 0.29 & 0.29 & 0.30 & 0.30 & 0.30 & 0.32 \\
\textbf{es} & 0.21 & 0.21 & 0.20 & 0.21 & 0.21 & 0.21 & 0.20 & 0.21 \\
\textbf{fr} & 0.25 & 0.27 & 0.27 & 0.27 & 0.27 & 0.27 & 0.28 & 0.29 \\
\textbf{ja} & 0.32 & 0.35 & 0.35 & 0.36 & 0.36 & 0.37 & 0.37 & 0.38 \\
\textbf{ko} & 0.15 & 0.17 & 0.17 & 0.17 & 0.17 & 0.17 & 0.17 & 0.17 \\
\textbf{pt} & 0.26 & 0.27 & 0.27 & 0.27 & 0.27 & 0.27 & 0.27 & 0.27 \\
\textbf{zh} & 0.27 & 0.27 & 0.27 & 0.29 & 0.28 & 0.28 & 0.28 & 0.29 \\
\midrule
\textbf{Avg.} & 0.25 & 0.26 & 0.26 & 0.26 & 0.27 & 0.27 & 0.26 & 0.27 \\
\bottomrule
\end{tabular}}
\caption{RougeL score calculated for base Aya 7B model instruction tuned on a single-task setting on [en, es, fr, ja, zh]. The columns represent the numbers of samples per non-English language used to finetune the model.}
\label{tab:10k-english_xlsum_single_7B}
\end{table}

\begin{table}[ht!]
\centering
\resizebox{0.5\textwidth}{!}{%
\begin{tabular}{l cccccccc}
\toprule
 & \textbf{0} & \textbf{100} & \textbf{200} & \textbf{400} & \textbf{800} & \textbf{1000} & \textbf{2000} & \textbf{5000} \\
\midrule
\textbf{ar} & 0.25 & 0.25 & 0.25 & 0.25 & 0.26 & 0.26 & 0.26 & 0.26 \\
\textbf{en} & 0.31 & 0.31 & 0.30 & 0.31 & 0.31 & 0.30 & 0.31 & 0.30 \\
\textbf{es} & 0.22 & 0.22 & 0.21 & 0.22 & 0.21 & 0.22 & 0.21 & 0.22 \\
\textbf{fr} & 0.26 & 0.27 & 0.27 & 0.27 & 0.27 & 0.29 & 0.29 & 0.29 \\
\textbf{ja} & 0.33 & 0.35 & 0.33 & 0.34 & 0.35 & 0.34 & 0.34 & 0.34 \\
\textbf{ko} & 0.19 & 0.18 & 0.18 & 0.19 & 0.18 & 0.19 & 0.18 & 0.19 \\
\textbf{pt} & 0.24 & 0.26 & 0.26 & 0.26 & 0.26 & 0.26 & 0.26 & 0.26 \\
\textbf{zh} & 0.27 & 0.28 & 0.27 & 0.28 & 0.28 & 0.28 & 0.28 & 0.28 \\
\midrule
\textbf{Avg.} & 0.26 & 0.26 & 0.26 & 0.26 & 0.27 & 0.27 & 0.27 & 0.27 \\
\bottomrule
\end{tabular}}
\caption{RougeL score calculated for base Aya 7B model instruction tuned on a single-task setting on [en, es, fr]. The columns represent the numbers of samples per non-English language used to finetune the model.}
\label{tab:latin-script_10k-english_xlsum_script-categories}
\end{table}

\begin{table}[ht!]
\centering
\resizebox{0.5\textwidth}{!}{%
\begin{tabular}{l cccccccc}
\toprule
 & \textbf{0} & \textbf{100} & \textbf{200} & \textbf{400} & \textbf{800} & \textbf{1000} & \textbf{2000} & \textbf{5000} \\
\midrule
\textbf{ar} & 0.24 & 0.24 & 0.24 & 0.24 & 0.25 & 0.25 & 0.25 & 0.25 \\
\textbf{en} & 0.32 & 0.30 & 0.32 & 0.32 & 0.31 & 0.31 & 0.31 & 0.30 \\
\textbf{es} & 0.22 & 0.22 & 0.22 & 0.22 & 0.22 & 0.21 & 0.22 & 0.22 \\
\textbf{fr} & 0.26 & 0.25 & 0.26 & 0.27 & 0.27 & 0.27 & 0.27 & 0.27 \\
\textbf{ja} & 0.32 & 0.34 & 0.35 & 0.36 & 0.36 & 0.37 & 0.37 & 0.38 \\
\textbf{ko} & 0.17 & 0.19 & 0.19 & 0.20 & 0.20 & 0.21 & 0.20 & 0.20 \\
\textbf{pt} & 0.25 & 0.25 & 0.26 & 0.26 & 0.26 & 0.26 & 0.27 & 0.26 \\
\textbf{zh} & 0.27 & 0.26 & 0.24 & 0.26 & 0.28 & 0.28 & 0.28 & 0.27 \\
\midrule
\textbf{Avg.} & 0.26 & 0.26 & 0.26 & 0.27 & 0.27 & 0.27 & 0.27 & 0.27 \\
\bottomrule
\end{tabular}}
\caption{RougeL score calculated for base Aya 7B model instruction tuned on a single-task setting on [en, ja, zh]. The columns represent the numbers of samples per non-English language used to finetune the model.}
\label{tab:non-latin-script_10k-english_xlsum_script}
\end{table}

\begin{table}[ht!]
\centering
\resizebox{0.5\textwidth}{!}{%
\begin{tabular}{l cccccccc}
\toprule
 & \textbf{0} & \textbf{100} & \textbf{200} & \textbf{400} & \textbf{800} & \textbf{1000} & \textbf{2000} & \textbf{5000} \\
\midrule
\textbf{ar} & 0.19 & 0.19 & 0.20 & 0.19 & 0.18 & 0.20 & 0.18 & 0.16 \\
\textbf{en} & 0.30 & 0.30 & 0.30 & 0.30 & 0.29 & 0.29 & 0.28 & 0.26 \\
\textbf{es} & 0.19 & 0.19 & 0.20 & 0.21 & 0.20 & 0.20 & 0.19 & 0.20 \\
\textbf{fr} & 0.24 & 0.26 & 0.25 & 0.26 & 0.26 & 0.26 & 0.25 & 0.25 \\
\textbf{ja} & 0.23 & 0.30 & 0.31 & 0.33 & 0.31 & 0.32 & 0.32 & 0.35 \\
\textbf{ko} & 0.11 & 0.12 & 0.12 & 0.12 & 0.12 & 0.11 & 0.10 & 0.09 \\
\textbf{pt} & 0.22 & 0.23 & 0.22 & 0.23 & 0.22 & 0.23 & 0.20 & 0.20 \\
\textbf{zh} & 0.23 & 0.24 & 0.25 & 0.26 & 0.27 & 0.26 & 0.23 & 0.23 \\
\midrule
\textbf{Avg.} & 0.21 & 0.23 & 0.23 & 0.24 & 0.23 & 0.23 & 0.22 & 0.22 \\
\bottomrule
\end{tabular}}
\caption{RougeL score calculated calculated for base Llama 3.1 8B instruction tuned on a single-task setting on [en, es, fr, ja, zh]. The columns represent the numbers of samples per non-English language used to finetune the model.}
\label{tab:10k-english_xlsum_llama31-8B_table}
\end{table}

\begin{table*}[ht!]
\centering
\resizebox{1.\textwidth}{!}{%
\begin{tabular}{l cccccccccccccccc}
\toprule
 & \textbf{0} & \textbf{100} & \textbf{500} & \textbf{1000} & \textbf{1500} & \textbf{2000} & \textbf{2500} & \textbf{3000} & \textbf{3500} & \textbf{4000} & \textbf{4500} & \textbf{5000} & \textbf{5500} & \textbf{6000} & \textbf{6500} & \textbf{7000} \\
\midrule
\textbf{ar} & 0.26 & 0.28 & 0.28 & 0.27 & 0.27 & 0.28 & 0.27 & 0.28 & 0.28 & 0.27 & 0.27 & 0.27 & 0.27 & 0.28 & 0.28 & 0.28 \\
\textbf{en} & 0.33 & 0.32 & 0.34 & 0.33 & 0.33 & 0.33 & 0.33 & 0.33 & 0.33 & 0.33 & 0.33 & 0.34 & 0.32 & 0.33 & 0.33 & 0.34 \\
\textbf{es} & 0.23 & 0.22 & 0.22 & 0.23 & 0.22 & 0.22 & 0.22 & 0.22 & 0.23 & 0.22 & 0.23 & 0.23 & 0.22 & 0.22 & 0.23 & 0.21 \\
\textbf{fr} & 0.29 & 0.28 & 0.28 & 0.31 & 0.31 & 0.31 & 0.31 & 0.32 & 0.32 & 0.31 & 0.31 & 0.31 & 0.31 & 0.31 & 0.31 & 0.31 \\
\textbf{ja} & 0.38 & 0.40 & 0.41 & 0.41 & 0.41 & 0.41 & 0.41 & 0.41 & 0.42 & 0.41 & 0.42 & 0.43 & 0.43 & 0.42 & 0.42 & 0.42 \\
\textbf{ko} & 0.20 & 0.22 & 0.20 & 0.21 & 0.22 & 0.22 & 0.21 & 0.22 & 0.23 & 0.21 & 0.21 & 0.21 & 0.21 & 0.22 & 0.21 & 0.21 \\
\textbf{pt} & 0.27 & 0.28 & 0.27 & 0.28 & 0.28 & 0.29 & 0.28 & 0.27 & 0.28 & 0.27 & 0.28 & 0.28 & 0.27 & 0.29 & 0.28 & 0.29 \\
\textbf{zh} & 0.26 & 0.31 & 0.31 & 0.32 & 0.32 & 0.33 & 0.33 & 0.31 & 0.31 & 0.31 & 0.31 & 0.31 & 0.31 & 0.31 & 0.32 & 0.32 \\
\midrule
\textbf{Avg.} & 0.28 & 0.29 & 0.29 & 0.30 & 0.30 & 0.30 & 0.30 & 0.30 & 0.30 & 0.29 & 0.29 & 0.30 & 0.29 & 0.30 & 0.30 & 0.30 \\
\bottomrule
\end{tabular}}
\caption{RougeL score calculated for base Aya 35B model instruction tuned on a multi-task setting on [en, es, fr, ja, zh]. The columns represent the numbers of samples per non-English language from XLSum used to finetune the model.}
\label{tab:10k-english_xlsum_multi_35B}
\end{table*}

\begin{table*}[ht!]
\centering
\resizebox{1.\textwidth}{!}{%
\begin{tabular}{l cccccccccccccccc}
\toprule
 & \textbf{0} & \textbf{100} & \textbf{500} & \textbf{1000} & \textbf{1500} & \textbf{2000} & \textbf{2500} & \textbf{3000} & \textbf{3500} & \textbf{4000} & \textbf{4500} & \textbf{5000} & \textbf{5500} & \textbf{6000} & \textbf{6500} & \textbf{7000} \\
\midrule
\textbf{ar} & 0.24 & 0.24 & 0.24 & 0.26 & 0.24 & 0.24 & 0.25 & 0.25 & 0.25 & 0.25 & 0.25 & 0.24 & 0.25 & 0.25 & 0.25 & 0.25 \\
\textbf{en} & 0.29 & 0.28 & 0.30 & 0.30 & 0.29 & 0.30 & 0.29 & 0.30 & 0.29 & 0.28 & 0.29 & 0.30 & 0.29 & 0.29 & 0.30 & 0.30 \\
\textbf{es} & 0.20 & 0.21 & 0.21 & 0.21 & 0.21 & 0.21 & 0.20 & 0.21 & 0.21 & 0.20 & 0.20 & 0.20 & 0.20 & 0.20 & 0.20 & 0.21 \\
\textbf{fr} & 0.25 & 0.26 & 0.27 & 0.27 & 0.26 & 0.25 & 0.27 & 0.27 & 0.27 & 0.27 & 0.27 & 0.28 & 0.27 & 0.27 & 0.27 & 0.27 \\
\textbf{ja} & 0.31 & 0.36 & 0.36 & 0.36 & 0.36 & 0.36 & 0.37 & 0.37 & 0.36 & 0.36 & 0.37 & 0.38 & 0.37 & 0.38 & 0.37 & 0.38 \\
\textbf{ko} & 0.14 & 0.17 & 0.18 & 0.17 & 0.18 & 0.18 & 0.18 & 0.18 & 0.18 & 0.17 & 0.17 & 0.17 & 0.17 & 0.17 & 0.16 & 0.17 \\
\textbf{pt} & 0.26 & 0.27 & 0.27 & 0.27 & 0.27 & 0.26 & 0.27 & 0.26 & 0.27 & 0.27 & 0.27 & 0.27 & 0.27 & 0.27 & 0.27 & 0.27 \\
\textbf{zh} & 0.26 & 0.27 & 0.27 & 0.29 & 0.28 & 0.28 & 0.29 & 0.28 & 0.29 & 0.29 & 0.29 & 0.29 & 0.29 & 0.28 & 0.28 & 0.29 \\
\midrule
\textbf{Avg.} & 0.25 & 0.26 & 0.26 & 0.26 & 0.26 & 0.26 & 0.26 & 0.26 & 0.26 & 0.26 & 0.26 & 0.27 & 0.26 & 0.26 & 0.26 & 0.26 \\
\bottomrule
\end{tabular}}
\caption{RougeL score calculated for base Aya 7B model instruction tuned on a multi-task setting on [en, es, fr, ja, zh]. The columns represent the numbers of samples per non-English language from XLSum used to finetune the model.}
\label{tab:10k-english_xlsum_multi_7B}
\end{table*}
\FloatBarrier

\subsection{Mathematical Reasoning}

Tables \ref{tab:10k-english_mcot_single_35B} through \ref{tab:10k-english_mcot_multi_7B} present the accuracy, calculated per language, on MGSM (8-shot) \citep{shi2022language} for all finetuned models.

\begin{table}[ht!]
\centering
\resizebox{0.5\textwidth}{!}{%
\begin{tabular}{l cccccccc}
\toprule
 & \textbf{0} & \textbf{100} & \textbf{200} & \textbf{400} & \textbf{800} & \textbf{1000} & \textbf{2000} & \textbf{5000} \\
\midrule
\textbf{ar} & 0.60 & 0.60 & 0.60 & 0.64 & 0.60 & 0.63 & 0.52 & 0.65 \\
\textbf{en} & 0.76 & 0.78 & 0.78 & 0.77 & 0.80 & 0.79 & 0.79 & 0.79 \\
\textbf{es} & 0.68 & 0.69 & 0.70 & 0.71 & 0.68 & 0.67 & 0.67 & 0.44 \\
\textbf{fr} & 0.66 & 0.65 & 0.64 & 0.65 & 0.66 & 0.64 & 0.68 & 0.64 \\
\textbf{ja} & 0.54 & 0.56 & 0.60 & 0.58 & 0.58 & 0.57 & 0.61 & 0.54 \\
\textbf{ko} & 0.50 & 0.51 & 0.52 & 0.53 & 0.59 & 0.51 & 0.60 & 0.60 \\
\textbf{pt} & 0.67 & 0.70 & 0.70 & 0.74 & 0.73 & 0.65 & 0.69 & 0.75 \\
\textbf{zh} & 0.56 & 0.66 & 0.60 & 0.64 & 0.66 & 0.66 & 0.66 & 0.67 \\
\midrule
\textbf{Avg.} & 0.62 & 0.64 & 0.64 & 0.66 & 0.66 & 0.64 & 0.65 & 0.64 \\
\bottomrule
\end{tabular}}
\caption{MGSM (8-shot) accuracy of base Aya 35B model instruction tuned on a single-task setting on [en, es, fr, ja, zh]. The columns represent the numbers of samples per non-English language used to finetune the model.}
\label{tab:10k-english_mcot_single_35B}
\end{table}

\begin{table}[ht!]
\centering
\resizebox{0.5\textwidth}{!}{%
\begin{tabular}{l cccccccc}
\toprule
 & \textbf{0} & \textbf{100} & \textbf{200} & \textbf{400} & \textbf{800} & \textbf{1000} & \textbf{2000} & \textbf{5000} \\
\midrule
\textbf{ar} & 0.01 & 0.02 & 0.06 & 0.10 & 0.09 & 0.10 & 0.13 & 0.17 \\
\textbf{en} & 0.50 & 0.56 & 0.54 & 0.51 & 0.54 & 0.57 & 0.55 & 0.56 \\
\textbf{es} & 0.42 & 0.44 & 0.47 & 0.52 & 0.54 & 0.54 & 0.49 & 0.55 \\
\textbf{fr} & 0.24 & 0.43 & 0.43 & 0.47 & 0.45 & 0.45 & 0.48 & 0.47 \\
\textbf{ja} & 0.07 & 0.12 & 0.18 & 0.25 & 0.24 & 0.26 & 0.26 & 0.35 \\
\textbf{ko} & 0.25 & 0.35 & 0.32 & 0.32 & 0.37 & 0.39 & 0.34 & 0.18 \\
\textbf{pt} & 0.34 & 0.46 & 0.47 & 0.46 & 0.47 & 0.50 & 0.50 & 0.51 \\
\textbf{zh} & 0.19 & 0.22 & 0.26 & 0.29 & 0.28 & 0.26 & 0.34 & 0.46 \\
\midrule
\textbf{Avg.} & 0.25 & 0.32 & 0.34 & 0.37 & 0.37 & 0.38 & 0.39 & 0.41 \\
\bottomrule
\end{tabular}}
\caption{MGSM (8-shot) accuracy of base Aya 7B model instruction tuned on a single-task setting on [en, es, fr, ja, zh]. The columns represent the numbers of samples per non-English language used to finetune the model.}
\label{tab:10k-english_mcot_single_7B}
\end{table}

\begin{table}[ht!]
\centering
\resizebox{0.5\textwidth}{!}{%
\begin{tabular}{l cccccccc}
\toprule
 & \textbf{0} & \textbf{100} & \textbf{200} & \textbf{400} & \textbf{800} & \textbf{1000} & \textbf{2000} & \textbf{5000} \\
\midrule
\textbf{ar} & 0.01 & 0.03 & 0.05 & 0.04 & 0.05 & 0.07 & 0.08 & 0.10 \\
\textbf{en} & 0.44 & 0.52 & 0.53 & 0.53 & 0.56 & 0.58 & 0.57 & 0.58 \\
\textbf{es} & 0.41 & 0.52 & 0.53 & 0.50 & 0.56 & 0.53 & 0.53 & 0.59 \\
\textbf{fr} & 0.21 & 0.41 & 0.47 & 0.42 & 0.48 & 0.44 & 0.45 & 0.47 \\
\textbf{ja} & 0.06 & 0.14 & 0.19 & 0.11 & 0.07 & 0.17 & 0.18 & 0.22 \\
\textbf{ko} & 0.26 & 0.31 & 0.34 & 0.36 & 0.36 & 0.34 & 0.38 & 0.35 \\
\textbf{pt} & 0.36 & 0.47 & 0.50 & 0.48 & 0.48 & 0.50 & 0.50 & 0.56 \\
\textbf{zh} & 0.22 & 0.22 & 0.26 & 0.27 & 0.26 & 0.28 & 0.29 & 0.33 \\
\midrule
\textbf{Avg.} & 0.25 & 0.33 & 0.36 & 0.34 & 0.35 & 0.36 & 0.37 & 0.40 \\
\bottomrule
\end{tabular}}
\caption{MGSM (8-shot) accuracy of base Aya 7B model instruction tuned on a single-task setting on [en, es, fr]. The columns represent the numbers of samples per non-English language used to finetune the model.}
\label{tab:latin-script_10k-english_mcot_script-categories}
\end{table}

\begin{table}[ht!]
\centering
\resizebox{0.5\textwidth}{!}{%
\begin{tabular}{l cccccccc}
\toprule
 & \textbf{0} & \textbf{100} & \textbf{200} & \textbf{400} & \textbf{800} & \textbf{1000} & \textbf{2000} & \textbf{5000} \\
\midrule
\textbf{ar} & 0.02 & 0.07 & 0.04 & 0.07 & 0.08 & 0.07 & 0.15 & 0.16 \\
\textbf{en} & 0.47 & 0.52 & 0.53 & 0.53 & 0.51 & 0.50 & 0.52 & 0.57 \\
\textbf{es} & 0.48 & 0.48 & 0.50 & 0.50 & 0.48 & 0.49 & 0.52 & 0.54 \\
\textbf{fr} & 0.26 & 0.33 & 0.30 & 0.33 & 0.36 & 0.38 & 0.39 & 0.46 \\
\textbf{ja} & 0.13 & 0.24 & 0.21 & 0.16 & 0.26 & 0.19 & 0.30 & 0.32 \\
\textbf{ko} & 0.29 & 0.34 & 0.37 & 0.37 & 0.37 & 0.36 & 0.36 & 0.14 \\
\textbf{pt} & 0.34 & 0.45 & 0.46 & 0.45 & 0.48 & 0.47 & 0.50 & 0.46 \\
\textbf{zh} & 0.23 & 0.25 & 0.25 & 0.28 & 0.37 & 0.33 & 0.34 & 0.34 \\
\midrule
\textbf{Avg.} & 0.28 & 0.34 & 0.33 & 0.34 & 0.36 & 0.35 & 0.38 & 0.37 \\
\bottomrule
\end{tabular}}
\caption{MGSM (8-shot) accuracy of base Aya 7B model instruction tuned on a single-task setting on [en, ja, zh]. The columns represent the numbers of samples per non-English language used to finetune the model.}
\label{tab:non-latin-script_10k-english_mcot_script}
\end{table}

\begin{table}[ht!]
\centering
\resizebox{0.5\textwidth}{!}{%
\begin{tabular}{l cccccccc}
\toprule
 & \textbf{0} & \textbf{100} & \textbf{200} & \textbf{400} & \textbf{800} & \textbf{1000} & \textbf{2000} & \textbf{5000} \\
\midrule
\textbf{ar} & 0.02 & 0.14 & 0.08 & 0.06 & 0.24 & 0.06 & 0.29 & 0.11 \\
\textbf{en} & 0.58 & 0.61 & 0.62 & 0.60 & 0.66 & 0.60 & 0.65 & 0.63 \\
\textbf{es} & 0.19 & 0.50 & 0.52 & 0.49 & 0.54 & 0.53 & 0.60 & 0.44 \\
\textbf{fr} & 0.16 & 0.47 & 0.46 & 0.49 & 0.52 & 0.51 & 0.54 & 0.52 \\
\textbf{ja} & 0.05 & 0.17 & 0.32 & 0.23 & 0.28 & 0.17 & 0.34 & 0.20 \\
\textbf{ko} & 0.12 & 0.34 & 0.28 & 0.32 & 0.31 & 0.34 & 0.28 & 0.28 \\
\textbf{pt} & 0.20 & 0.54 & 0.47 & 0.52 & 0.55 & 0.51 & 0.31 & 0.56 \\
\textbf{zh} & 0.17 & 0.48 & 0.47 & 0.48 & 0.47 & 0.38 & 0.45 & 0.46 \\
\midrule
\textbf{Avg.} & 0.19 & 0.40 & 0.40 & 0.40 & 0.45 & 0.39 & 0.43 & 0.40 \\
\bottomrule
\end{tabular}}
\caption{MGSM (8-shot) accuracy of base Llama 3.1 8B model instruction tuned on a single-task setting on [en, es, fr, ja, zh]. The columns represent the numbers of samples per non-English language used to finetune the model.}
\label{tab:10k-english_mcot_llama31-8B_table}
\end{table}

\begin{table*}[ht!]
\centering
\resizebox{1.\textwidth}{!}{%
\begin{tabular}{l cccccccccccccccc}
\toprule
 & \textbf{0} & \textbf{100} & \textbf{500} & \textbf{1000} & \textbf{1500} & \textbf{2000} & \textbf{2500} & \textbf{3000} & \textbf{3500} & \textbf{4000} & \textbf{4500} & \textbf{5000} & \textbf{5500} & \textbf{6000} & \textbf{6500} & \textbf{7000} \\
\midrule
\textbf{ar} & 0.60 & 0.60 & 0.63 & 0.62 & 0.66 & 0.65 & 0.64 & 0.69 & 0.68 & 0.68 & 0.66 & 0.67 & 0.66 & 0.68 & 0.65 & 0.67 \\
\textbf{en} & 0.76 & 0.74 & 0.78 & 0.77 & 0.78 & 0.78 & 0.75 & 0.78 & 0.75 & 0.74 & 0.82 & 0.76 & 0.78 & 0.76 & 0.75 & 0.76 \\
\textbf{es} & 0.27 & 0.42 & 0.47 & 0.68 & 0.70 & 0.60 & 0.59 & 0.71 & 0.53 & 0.72 & 0.72 & 0.62 & 0.71 & 0.61 & 0.70 & 0.73 \\
\textbf{fr} & 0.63 & 0.65 & 0.65 & 0.65 & 0.69 & 0.62 & 0.68 & 0.67 & 0.68 & 0.66 & 0.64 & 0.64 & 0.64 & 0.65 & 0.66 & 0.68 \\
\textbf{ja} & 0.61 & 0.61 & 0.58 & 0.59 & 0.60 & 0.58 & 0.60 & 0.55 & 0.58 & 0.63 & 0.60 & 0.61 & 0.59 & 0.62 & 0.59 & 0.62 \\
\textbf{ko} & 0.54 & 0.58 & 0.57 & 0.59 & 0.60 & 0.60 & 0.60 & 0.59 & 0.59 & 0.60 & 0.63 & 0.60 & 0.60 & 0.64 & 0.60 & 0.60 \\
\textbf{pt} & 0.68 & 0.68 & 0.69 & 0.72 & 0.72 & 0.74 & 0.70 & 0.72 & 0.68 & 0.74 & 0.73 & 0.75 & 0.73 & 0.77 & 0.76 & 0.72 \\
\textbf{zh} & 0.63 & 0.68 & 0.67 & 0.66 & 0.68 & 0.67 & 0.68 & 0.69 & 0.72 & 0.71 & 0.71 & 0.71 & 0.72 & 0.72 & 0.71 & 0.70 \\
\midrule
\textbf{Avg.} & 0.59 & 0.62 & 0.63 & 0.66 & 0.68 & 0.65 & 0.66 & 0.67 & 0.65 & 0.69 & 0.69 & 0.67 & 0.68 & 0.68 & 0.68 & 0.69 \\
\bottomrule
\end{tabular}}
\caption{MGSM (8-shot) accuracy of base Aya 35B model instruction tuned on a multi-task setting on [en, es, fr, ja, zh]. The columns represent the numbers of samples per non-English language from mCoT-math used to finetune the model.}
\label{tab:10k-english_mcot_multi_35B}
\end{table*}

\begin{table*}[ht!]
\centering
\resizebox{1.\textwidth}{!}{%
\begin{tabular}{l cccccccccccccccc}
\toprule
 & \textbf{0} & \textbf{100} & \textbf{500} & \textbf{1000} & \textbf{1500} & \textbf{2000} & \textbf{2500} & \textbf{3000} & \textbf{3500} & \textbf{4000} & \textbf{4500} & \textbf{5000} & \textbf{5500} & \textbf{6000} & \textbf{6500} & \textbf{7000} \\
\midrule
\textbf{ar} & 0.00 & 0.00 & 0.00 & 0.00 & 0.02 & 0.00 & 0.01 & 0.01 & 0.01 & 0.02 & 0.02 & 0.02 & 0.03 & 0.02 & 0.00 & 0.02 \\
\textbf{en} & 0.49 & 0.44 & 0.30 & 0.44 & 0.46 & 0.46 & 0.40 & 0.38 & 0.40 & 0.44 & 0.49 & 0.47 & 0.54 & 0.52 & 0.47 & 0.61 \\
\textbf{es} & 0.36 & 0.42 & 0.20 & 0.29 & 0.26 & 0.21 & 0.20 & 0.22 & 0.19 & 0.18 & 0.14 & 0.22 & 0.26 & 0.26 & 0.23 & 0.27 \\
\textbf{fr} & 0.19 & 0.37 & 0.11 & 0.27 & 0.38 & 0.27 & 0.22 & 0.27 & 0.16 & 0.30 & 0.22 & 0.30 & 0.32 & 0.30 & 0.38 & 0.44 \\
\textbf{ja} & 0.18 & 0.20 & 0.11 & 0.18 & 0.27 & 0.25 & 0.24 & 0.24 & 0.19 & 0.30 & 0.32 & 0.30 & 0.36 & 0.28 & 0.22 & 0.32 \\
\textbf{ko} & 0.26 & 0.27 & 0.14 & 0.24 & 0.20 & 0.23 & 0.23 & 0.16 & 0.36 & 0.34 & 0.24 & 0.28 & 0.29 & 0.22 & 0.10 & 0.22 \\
\textbf{pt} & 0.12 & 0.07 & 0.01 & 0.08 & 0.28 & 0.18 & 0.28 & 0.15 & 0.23 & 0.14 & 0.21 & 0.34 & 0.33 & 0.34 & 0.03 & 0.48 \\
\textbf{zh} & 0.23 & 0.26 & 0.12 & 0.20 & 0.34 & 0.34 & 0.28 & 0.30 & 0.30 & 0.33 & 0.31 & 0.26 & 0.32 & 0.33 & 0.29 & 0.35 \\
\midrule
\textbf{Avg.} & 0.23 & 0.26 & 0.12 & 0.21 & 0.28 & 0.24 & 0.23 & 0.22 & 0.23 & 0.26 & 0.24 & 0.27 & 0.31 & 0.28 & 0.22 & 0.34 \\
\bottomrule
\end{tabular}}
\caption{MGSM (8-shot) accuracy of base Aya 7B model instruction tuned on a multi-task setting on [en, es, fr, ja, zh]. The columns represent the numbers of samples per non-English language from mCoT-math used to finetune the model.}
\label{tab:10k-english_mcot_multi_7B}
\end{table*}
\FloatBarrier

\section{Additional Plots}

\begin{figure}[ht!]
  \centering
  \includegraphics[width=\linewidth]{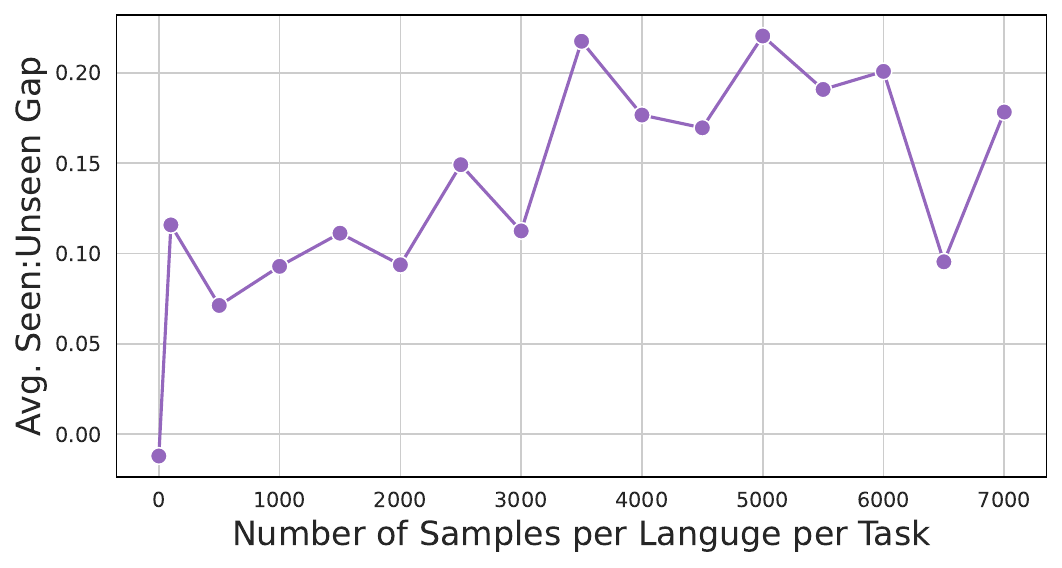}
  \caption{Plot showing the increasing gap considering average performance on Dolly200 (ratings) between seen and unseen languages in the multi-task finetuning setting for the Aya 7B base model.}
  \label{fig:seen_unseen_gap}
\end{figure}

\end{document}